\documentclass{article}

\usepackage{microtype}
\usepackage{graphicx}
\usepackage{subcaption}
\usepackage{booktabs} %

\usepackage{hyperref}

\usepackage[accepted]{icml2024}

\usepackage{amsmath}
\usepackage{amssymb}
\usepackage{mathtools}
\usepackage{amsthm}

\usepackage{tikz} %
\usepackage{multirow}
\usepackage{url}
\usepackage{graphicx}
\usepackage{wrapfig}
\usepackage{subcaption}
\usepackage{tcolorbox}
\usepackage{mathtools}
\usepackage{adjustbox}
\usepackage{subcaption}
\usepackage{algorithm}
\usepackage{selectp}
\usepackage{xcolor}

\usepackage[capitalize,noabbrev]{cleveref}

\theoremstyle{plain}

\theoremstyle{definition}

\theoremstyle{remark}

\renewcommand{\algorithmicrequire}{\textbf{Input:}}
\renewcommand{\algorithmicensure}{\textbf{Output:}}

\DeclarePairedDelimiter\ceil{\lceil}{\rceil}

\def\nocheckmark{\tikz\draw[scale=0.4,fill=red,draw=red](.25,.1) -- (1,.7) -- (.25,.15) -- cycle (0.75,0.15) -- (0.77,0.15)  -- (0.3,0.7) -- cycle;}
\def\halfcheckmark{\tikz\draw[scale=0.4,fill=orange,draw=orange](0,.35) -- (.25,0) -- (1,.7) -- (.25,.15) -- cycle (0.75,0.15) -- (0.78,0.15)  -- (0.5,0.7) -- cycle;}
\def\checkmark{\tikz\draw[scale=0.4,fill=green,draw=green](0,.35) -- (.25,0) -- (1,.7) -- (.25,.15) -- cycle;}

\icmltitlerunning{Memory Efficient Neural Processes via Constant Memory Attention Block}

\begin{document}

\twocolumn[
\icmltitle{Memory Efficient Neural Processes via Constant Memory Attention Block}

\icmlsetsymbol{equal}{*}

\begin{icmlauthorlist}
\icmlauthor{Leo Feng}{sch,comp}
\icmlauthor{Frederick Tung}{comp}
\icmlauthor{Hossein Hajimirsadeghi}{comp}
\icmlauthor{Yoshua Bengio}{sch}
\icmlauthor{Mohamed Osama Ahmed}{comp}
\end{icmlauthorlist}

\icmlaffiliation{comp}{Borealis AI, Canada}
\icmlaffiliation{sch}{Mila -- Université de Montréal, Canada}

\icmlcorrespondingauthor{Leo Feng}{leo.feng@mila.quebec}

\icmlkeywords{Machine Learning, ICML, Attention, Neural Processes, Meta-learning, Efficiency}

\vskip 0.3in
]

\printAffiliationsAndNotice{\icmlEqualContribution} 

\begin{abstract}
    Neural Processes (NPs) are popular meta-learning methods for efficiently modelling predictive uncertainty.
    Recent state-of-the-art methods, however, leverage expensive attention mechanisms, limiting their applications, particularly in low-resource settings. 
    In this work, we propose Constant Memory Attentive Neural Processes (CMANPs), an NP variant that only requires \textbf{constant} memory. To do so, we first propose an efficient update operation for Cross Attention. 
    Leveraging the update operation, we propose Constant Memory Attention Block (CMAB), a novel attention block that (i) is permutation invariant, (ii) computes its output in constant memory, and (iii) performs constant computation updates. 
    Finally, building on CMAB, we detail Constant Memory Attentive Neural Processes. 
    Empirically, we show CMANPs achieve state-of-the-art results on popular NP benchmarks while being significantly more memory efficient than prior methods. 
\end{abstract}

\section{Introduction}\label{sec:intro}

Neural Processes (NPs) are a popular family of meta-learning models for uncertainty estimation.
They are particularly useful in low-resource settings due to not requiring retraining from scratch given new data. 
Over recent years, NPs have been applied to a wide variety of settings such as graph link prediction~\citep{liang2022neural}, continual learning~\citep{requeima2019fast}, and recommender systems~\citep{lin2021task}.
With the growing popularity of low-memory/compute domains (e.g., IoT devices and mobile phones), deployed models in these low-resource settings must be memory efficient. Furthermore, memory access is energy intensive~\citep{zhi2022memory}. 
As such, the design of memory-efficient models is crucial for settings with a limited energy supply (e.g., battery-powered devices such as mobile robots).

However, the state-of-the-art NP methods~\citep{nguyen2022transformer, feng2023latent} are attention-based methods that require substantial amounts of memory, limiting their applicability. 
For example, Transformer Neural Processes (TNPs)~\citep{nguyen2022transformer} leverage Transformers~\citep{vaswani2017attention}. As such, TNPs are computationally expensive, scaling quadratically with the size of the context and query dataset. Latent Bottlenecked Attentive Neural Processes (LBANPs)~\citep{feng2023latent} leverage Perceiver's iterative attention~\citep{jaegle2021perceiver} and require $\mathcal{O}(k |\mathcal{D}_C|)$ memory where $|\mathcal{D}_C|$ is the size of the context dataset and $k$ is a hyperparameter that scales with the difficulty of the task and the context dataset size.

Tackling this issue, we propose Constant Memory Attentive Neural Processes (CMANPs), a novel attention-based NP that is competitive with prior state-of-the-art while only requiring a constant amount of memory. To do so, we first (1) improve the computational efficiency of Cross Attention and (2) propose a novel efficient attention block.

More specifically, we propose (1) an exact update operation for Cross Attention that allows it to be efficiently updated with new context data (Section \ref{sec:method:cross_att:efficient_update_property}). 
Using the efficient updates operation, we show that Cross Attention can also be computed memory-efficiently (Section \ref{sec:method:cross_att:reduce_memory}). 
Leveraging the aforementioned efficiency properties of Cross Attention, we propose (2) Constant Memory Attention Block (CMAB) (Section \ref{sec:method:cmab}), a novel attention block, that (i) is permutation invariant, (ii) computes its output in constant memory, and (iii) performs updates in constant computation. 

Finally, building on CMABs, we propose Constant Memory Attentive Neural Processes (CMANPs) (Section \ref{sec:method:CMANPs}). By using CMABs, CMANPs (i) only require constant memory, making it naturally scalable in the number of data points and (ii) allow for updates to the context to be performed efficiently unlike prior state-of-the-art NPs. Leveraging the efficient updates property, we further introduce an Autoregressive Not-Diagonal extension (Section \ref{sec:method:CMANP_AND}) which only requires constant memory unlike the quadratic memory required by all prior Not-Diagonal extensions of NPs. In the experiments, CMANPs achieve state-of-the-art results on popular NP benchmarks while being significantly more memory efficient than prior state-of-the-art methods.

\section{Background} \label{sec:background}

\subsection{Meta-learning for Predictive Uncertainty Estimation}

In meta-learning for predictive uncertainty estimation, models are trained on a distribution of tasks $\Omega(\mathcal{T})$ to model a probabilistic predictive distribution.
A task $\mathcal{T}$ is a tuple ($\mathcal{X}$, $\mathcal{Y}$, $\mathcal{L}$, $q$) where $\mathcal{X}, \mathcal{Y}$ are the input and output space respectively, $\mathcal{L}$ is the task-specific loss function, and $q(x, y)$ is the task-specific distribution over data points. 
During each meta-training iteration, $B$ tasks $\mathbf{T} = \{\mathcal{T}_i\}_{i=1}^{B}$ are sampled from the task distribution $\Omega(\mathcal{T})$. 
For each task $\mathcal{T}_{i} \in \mathbf{T}$, a context dataset $\mathcal{D}_C^i = \{ (x, y)^{i,j}\}_{j=1}^{N}$ and a target dataset $\mathcal{D}_T^i= \{ (x, y)^{i, j}\}_{j=1}^{M}$ are sampled from the task-specific data point distribution $q_{\mathcal{T}_{i}}$. $N$ is a fixed number of context data points and $M$ is a fixed number of target data points. 
 The model is adapted using the context dataset. Afterwards, the target dataset is used to evaluate the effectiveness of the adaptation and adjust the adaptation rule accordingly.

\subsection{Neural Processes}
Neural Processes (NPs)~\citep{garnelo2018neural} are meta-learned models that define a family of conditional distributions. 
Specifically, NPs condition on an arbitrary amount of context data points (labelled data points) and make predictions for a batch of target data points, while preserving invariance in the ordering of the context dataset. 
NPs consist of three phases: conditioning, querying, and updating.
However, recent state-of-the-art NP models~\citep{nguyen2022transformer, feng2023latent} leverage attention, causing these phases to require large amounts of memory and limiting their applicability.

\textbf{Conditioning:} In the conditioning phase, the model encodes the context dataset $\mathcal{D}_{C}$. Neural Processes model functional uncertainty by encoding the context dataset as a Gaussian latent variable: $z_C \sim q(z | \mathcal{D}_{C})$ where $q(z | \mathcal{D}_{C}) = \mathcal{N}(z; \mu_C, \sigma_C^2)$ and $\mu_C, \sigma_C = f_{encoder}(\mathcal{D}_{C})$. 
Conditional variants~\citep{garnelo2018conditional} instead compute a deterministic encoding, i.e., $z_C = f_{encoder}(\mathcal{D}_{C})$.

\textbf{Querying:} In the querying phase, given target data points $x_{T}$ to make predictions for, the NP models the predictive distribution $p(y_{T} | x_{T}, z_C)$.

\textbf{Updating:} In the updating phase, the model receives new context data points $\mathcal{D}_{U}$ and uses it to compute new encodings, i.e., re-computing $z_C$ given $\mathcal{D}_{C} \gets \mathcal{D}_{C} \cup \mathcal{D}_{U}$.

During training, deterministic NP variants maximize the log-likelihood directly. In contrast, stochastic variants maximize an evidence lower bound (ELBO) of the log-likelihood:
\begin{align*}
    \log p(y_{T} | x_{T}, \mathcal{D}_{C}) \geq &\mathbb{E}_{q(z | \mathcal{D}_{C} \cup \mathcal{D}_{T})}\left[\log p(y_{T} | x_{T}, z)\right]  \\
    &- \mathrm{KL}(q(z | \mathcal{D}_{C} \cup \mathcal{D}_{T}) || p(z | \mathcal{D}_{C}))
\end{align*}

\section{Methodology} \label{sec:method}

In this section, we propose Constant Memory Attentive Neural Processes (CMANPs), a novel attention-based NP that only requires constant memory for the conditioning, querying, and updating phases. To do so, we begin by proposing an efficient update operation for Cross Attention (Section \ref{sec:method:cross_att}). Leveraging the update operation, we propose Constant Memory Attention Block (CMAB) (Section \ref{sec:method:cmab}), a memory-efficient attention block. Finally, building on CMAB, we propose Constant Memory Attentive Neural Processes (Section \ref{sec:method:CMANPs}).

\subsection{Cross Attention} \label{sec:method:cross_att}

Cross Attention (CA) is widely used in state-of-the-art Neural Processes during the conditioning, querying, and updating phases and makes up a significant portion of their computational complexity. 
Cross Attention retrieves information from a set of $\mathcal{D}_C$ context tokens for a given set of query tokens $L$ via a weighted average as follows:
$$\mathrm{CrossAttention}(L, \mathcal{D}_C) = \mathrm{softmax}(Q K^T)V
$$
where $Q = L W_q$ is the query matrix, $K = \mathcal{D}_C W_k$ is the key matrix, and $V = \mathcal{D}_C W_v$ is the value matrix. $W_q, W_k, W_v \in \mathbb{R}^{d \times d}$ are weight matrices (learned parameters) that project the input tokens. $\mathrm{softmax}(QK^T)$ computes the weight for each context token in the weighted average. 

When given a new set of context tokens $\mathcal{D}_U$, computing the updated output $\mathrm{CrossAttention}(L, \mathcal{D}_C \cup \mathcal{D}_U)$ from $\mathrm{CrossAttention}(L, \mathcal{D}_C)$ requires $\mathcal{O}((|\mathcal{D}_C| + |\mathcal{D}_U|)|L|)$ memory and computation. In practice, $|\mathcal{D}_C|$ is significantly larger than $|\mathcal{D}_U|$ and $|L|$, making the complexity's reliance on $|\mathcal{D}_C|$ a limitation. 

In this section, we propose an efficient update operation that computes $\mathrm{CrossAttention}(L, \mathcal{D}_C \cup \mathcal{D}_U)$ from $\mathrm{CrossAttention}(L, \mathcal{D}_C)$ in only $\mathcal{O}(|\mathcal{D}_U||L|)$ computation: 
\begin{align*}
    \mathrm{CrossAttention}&(L, \mathcal{D}_C\, \cup\, \mathcal{D}_U) = \\
    &\mathrm{UPDATE}(\mathcal{D}_U, \mathrm{CrossAttention}(L, \mathcal{D}_C))
\end{align*}
Leveraging this efficient update operation, we then show that $\mathrm{CrossAttention}(L, \mathcal{D}_C)$ can be efficiently computed in only $\mathcal{O}(|L|)$ memory. Notably, these complexities are independent of $\mathcal{D}_C$ making them computationally efficient and naturally scalable with respect to the number of context tokens $|\mathcal{D}_C|$. A formal proof and derivation of the efficient update operation is included in Appendix \ref{appendix:proof:efficient_updates}.

\subsubsection{Efficient Update Property} \label{sec:method:cross_att:efficient_update_property}

When computing the updated value of $\mathrm{CA'} = \mathrm{CrossAttention}(L, \mathcal{D}_C\, \cup\, \mathcal{D}_U)$ with a new set of $\mathcal{D}_U$ context tokens, $|\mathcal{D}_U|$ rows are added to the key, value matrices. Since the attention weights are computed via products between the keys (with $|\mathcal{D}_U|$ added rows) and queries (unchanged), thus the new output can be computed from $\mathrm{CA} = \mathrm{CrossAttention}(L, \mathcal{D}_C)$ via a rolling average in $\mathcal{O}(|\mathcal{D}_U| |L|)$.
An implementation of the proposed $\mathrm{UPDATE}$ operation for $j \in \{1, 2, \ldots, |L| \}$ is as follows:
$$
    \mathrm{CA}_j' = \frac{c_j}{c'_j} \times \mathrm{CA}_j \,\,+ \sum_{i=|\mathcal{D}_C|+1}^{|\mathcal{D}_C|+|\mathcal{D}_U|} \frac{e^{s_i}}{c'_j} v_i
$$
where $s_i = Q_{j,:}(K_{i,:})^T$, $v_i = V_{i, :}$, and $c'_j = c_j + \sum_{i=|\mathcal{D}_C|+1}^{|\mathcal{D}_C|+|\mathcal{D}_U|} \exp(s_i)$. $c'$ are normalizing constants for $\mathrm{CA}'$ that are computed via a rolling sum from cached normalizing constants $c$. 
When computing the updated value $\mathrm{CA}'$, $c'$ replaces $c$ as the cached normalizing constants. 
Computing $\mathrm{CA}_j'$ and $c'_j$ via these rolling average and summation thus only requires $\mathcal{O}(|\mathcal{D}_U|)$ operations when given $\mathrm{CA}_j$ and $c_j$. 
Since there are $|L|$ values for $j$, the total amount of computation is thus $\mathcal{O}(|\mathcal{D}_U||L|)$. 
In practice, however, this is not a stable implementation. The computation of $c_j' = c_j + \sum_{i=|\mathcal{D}_C|+1}^{|\mathcal{D}_C|+|\mathcal{D}_U|} \exp(s_i)$ can quickly run into numerical issues such as underflow and overflow due to being a sum of exponentials.

As such, instead of computing $c'$ in the update, we propose to compute $\log(c')$, resulting in the following update:
\begin{align*}
    \mathrm{CA'}_j = \exp(\log(c_j) - &\log(c'_j)) \times \mathrm{CA}_j + \\
    &\sum_{i=|\mathcal{D}_C|+1}^{|\mathcal{D}_C|+|\mathcal{D}_U|} \exp(s_i - \log(c'_j)) v_i
\end{align*}
where $\log(c'_j) = \log(c_j) + \mathrm{softplus}(t_j)$ and $t_j = \log(\sum_{i=|\mathcal{D}_C|+1}^{|\mathcal{D}_C| + |\mathcal{D}_U|} \exp(s_i - \log(c)))$. $t_j$ can be computed accurately using the log-sum-exp trick, avoiding the underflow and overflow issues.

\subsubsection{Reducing Memory Usage} \label{sec:method:cross_att:reduce_memory}

A naive computation of $\mathrm{CrossAttention}(L, \mathcal{D}_C)$ would require linear memory complexity in the number of context tokens, i.e., $\mathcal{O}(|\mathcal{D}_C| |L|)$. Instead, we propose to leverage the aforementioned update operation and split the input data $\mathcal{D}_C$ into $|\mathcal{D}_C|/b_C$ batches of input data points of size up to $b_C$ (a pre-specified constant), i.e., $\mathcal{D}_C = \cup_{i=1}^{|\mathcal{D}_C|/b_C} \mathcal{D}_C^{(i)}$. Computing $\mathrm{CrossAttention}(L, \mathcal{D}_C)$ is then equivalent to performing an update $|\mathcal{D}_C|/b_C - 1$ times:
\begin{align*}
    \mathrm{CA}&(L, \mathcal{D}_C) = \mathrm{UPDATE}(\mathcal{D}_C^{(1)}, 
        \mathrm{UPDATE}(
            \mathcal{D}_C^{(2)}, \ldots \\
            &\mathrm{UPDATE}(
                \mathcal{D}_C^{(|\mathcal{D}_C|/b_C-1)}, \mathrm{CA}(L_B, \mathcal{D}_C^{(|\mathcal{D}_C|/b_C}))
            )
        )
    )
\end{align*}
Computing $\mathrm{CrossAttention}(L,\mathcal{D}_C^{(|\mathcal{D}_C|/b_C)})$ requires $\mathcal{O}(b_C|L|)$ memory. After its computation, the memory can be freed up, so that each of the subsequent $\mathrm{UPDATE}$ operations can re-use the memory space. Each of the update operations also only uses $\mathcal{O}(b_C|L|)$ constant memory. As such, $\mathrm{CrossAttention}(L, \mathcal{D}_C)$ only needs $\mathcal{O}(b_C |L|) = \mathcal{O}(|L|)$ memory in total, i.e., a memory amount that is independent of the number of context data points. Notably, $b_C$ is a hyperparameter constant which trades off the memory and time complexity.

\subsection{Constant Memory Attention Block (CMAB)} \label{sec:method:cmab}

\begin{figure*}[h]
    \centering
    \includegraphics[width=0.8\linewidth]{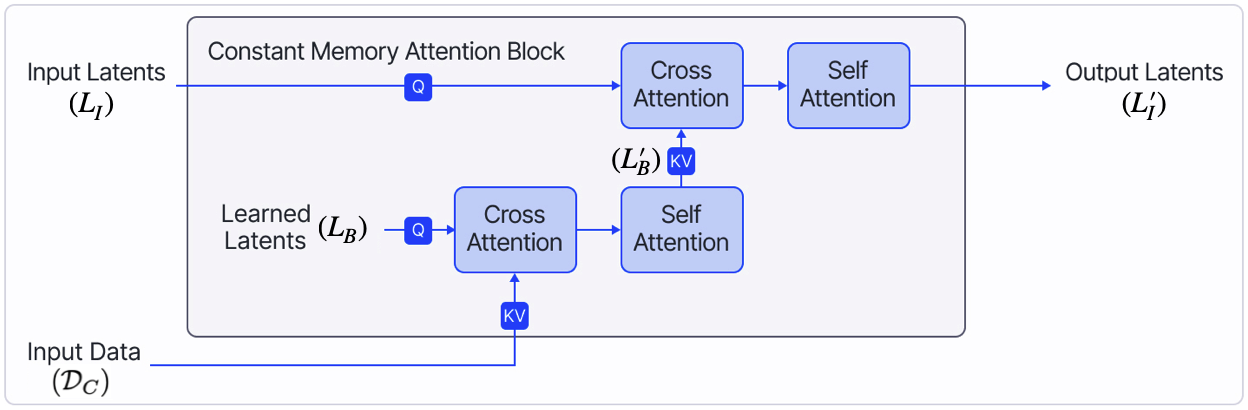}
    \caption{Constant Memory Attention Block (CMAB).
    CMABs are stackable attention blocks that (i) are permutation invariant in the input data, (ii) compute their output in constant memory, and (iii) compute updates in constant computation per new data point. Notably, $L_B$ is a learned set of latents unique to the block, allowing CMABs to take advantage of the efficient update property of Cross Attention. 
    }
    \label{fig:CMAB}
\end{figure*}

Efficiently computing $\mathrm{CrossAttention}(L, \mathcal{D}_C\, \cup\, \mathcal{D}_U)$ from $\mathrm{CrossAttention}(L, \mathcal{D}_C)$ require (1) the query tokens remain the same and (2) the new tokens being appended to the old context tokens.
However, these requirements do not hold for prior attention models~\citep{vaswani2017attention,lee2019set,jaegle2021perceiver}.
When these attention blocks are stacked, the query tokens of later attention blocks end up being conditioned on the context dataset\footnote{Further details are included in Appendix \ref{appendix:proof:eff_upd_perceiver}.}. As such, when the context dataset changes, the query tokens of these later attention blocks change as well, meaning that the efficient updates property of Cross Attention cannot be applied to existing stacked attention models such as Perceiver~\citep{jaegle2021perceiverio, jaegle2021perceiver}. 
As such, in this section, we specially design the Constant Memory Attention Block (CMAB) that is capable of leveraging the efficiency properties of cross attention while being stacked\footnote{See Appendix \ref{appendix:cmab_intuition} for insights regarding the construction of CMAB.}.

The Constant Memory Attention Block (Figure \ref{fig:CMAB}) takes as input a fixed-sized set of latent vectors $L_I$ and the context data $\mathcal{D}_C$ and outputs a new set of latent vectors $L_I'$. Within each CMAB is a unique fixed-sized set of latents $L_B$ learned during training. When stacking CMABs, the output latent vectors of the previous CMAB are fed as the input latent vectors to the next, i.e., $L_I \gets L_I'$. The value of $L_I$ of the first stacked CMAB block is learned.

In the first phase, CMAB applies cross attention between the context data $\mathcal{D}_C$ and the block-wise latent vectors $L_B$, compressing the data representation. Crucially, since $L_B$ is a set of fixed learned latent vectors unique to each CMAB block, the query tokens are fixed as required regardless of stacking; as a result, this cross attention possesses the efficiency properties of Cross Attention.
Afterwards, a self attention is applied to compute higher-order information:
\begin{align*}
    L_B' = \mathrm{SelfAttention}(\mathrm{CrossAttention}(L_B, \mathcal{D}_C))
\end{align*} 

The second phase is designed to make CMABs stackable, i.e., computing $L_I'$ from $L_I$. Specifically, cross attention is performed by taking in as query tokens the prior latents $L_I$ and as context the compressed data representation $L_B'$. Afterwards, an additional self-attention is used to compute further higher-order information, resulting in the output vectors $L_I'$:
\begin{align*}
    L_I' &= \mathrm{SelfAttention}(\mathrm{CrossAttention}(L_I, L_B'))
\end{align*} 

In terms of the computational complexity of CMAB, the first phase has an overall complexity of $\mathcal{O}(|\mathcal{D}_C| |L_B| + |L_B|^2)$ and the second phase has an overall complexity of $\mathcal{O}(|L_B| |L_I| + |L_I|^2)$. Since the number of latents $|L_B|$ and $|L_I|$ are hyperparameter constants, therefore these complexities are all constant except for the $\mathcal{O}(|\mathcal{D}_C||L_B|)$ factor induced by the first phase's $\mathrm{CrossAttention}(L_B, \mathcal{D}_C)$. However, due to $L_B$ being fixed, CMABs can leverage the efficiency properties of Cross Attention, resulting in the following CMAB-specific properties: (1) perform updates in constant computation and (2) compute their output in constant memory. In contrast, methods like Transformer or Perceiver require linear or quadratic memory and re-computing from scratch when given new context data.
Furthermore, since Cross Attention is permutation invariant in the context, CMABs are also (3) permutation invariant by default.

\subsection{Constant Memory Attentive Neural Processes (CMANPs)} \label{sec:method:CMANPs}

\begin{figure*}[h]
    \centering
    \includegraphics[width=0.85\textwidth]{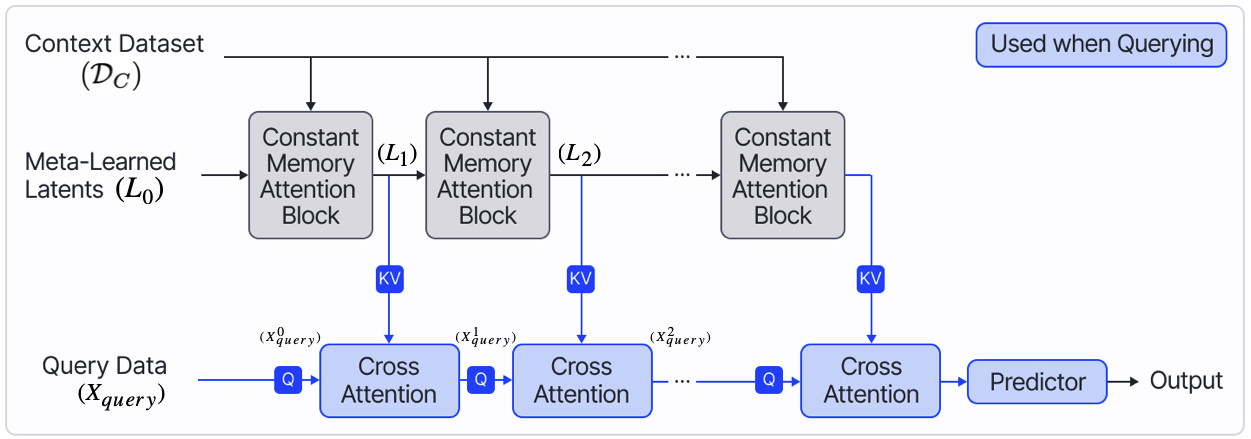}
    \caption{Constant Memory Attentive Neural Processes (CMANPs). CMANPs perform the conditioning, querying, and updating phases in constant memory with respect to the size of the context dataset $|\mathcal{D}_C|$.}
    \label{fig:CMANPs}
\end{figure*}

In this section, we introduce Constant Memory Attentive Neural Processes (CMANPs), a memory efficient variant of Neural Processes (Figure \ref{fig:CMANPs}) that leverages the stacked CMAB blocks for efficiency.
The conditioning, querying, and updating phases in CMANPs work as follows:

\textbf{Conditioning Phase:} In the conditioning phase, the CMAB blocks encode the context dataset into a set of latent vectors $L_i$.
The first block takes as input a set of meta-learned latent vectors $L_0$ (i.e., $L_I$ in CMABs) and the context dataset $\mathcal{D}_C$, and outputs a set of encodings $L_1$ (i.e., $L_I'$ in CMABs).  
The output latents of each block are passed as the input latents to the next CMAB block. 
\begin{align*}
    L_i &= \mathbf{CMAB}(L_{i-1}, \mathcal{D}_C)
\end{align*}
Since CMABs can compute their output in constant memory, CMANPs can also perform this conditioning phase in constant memory. 

\textbf{Querying Phase:} In the querying phase, the deployed model retrieves information from the fixed size outputs of the CMAB blocks ($L_i$) to make predictions for the query data points ($X_{query}$). 
Beginning with $X_{query}^0 \gets X_{query}$, when making a prediction for the query data points $X_{query}$, information is retrieved via cross-attention. 
\begin{align*}
    X^i_{query} &= \mathrm{CrossAttention}(X^{i-1}_{query}, L_{i}) \\
\end{align*}
\textbf{Updating Phase:} In the updating phase, the NP receives a batch of new data points $\mathcal{D}_U$ to include in the context dataset. 
CMANPs leverage the efficient update mechanism of CMABs to perform this phase efficiently (i.e., constant computation per new data point).
Beginning with $L^{updated}_0 \gets L_0$, the CMAB blocks are updated sequentially using the updated output of the previous CMAB block as follows:
\begin{align*}
    L^{updated}_i &= \mathbf{CMAB}(L^{updated}_{i-1}, \mathcal{D}_C\,\cup\,\mathcal{D}_U)
\end{align*}
Since CMABs perform updates in constant memory irrespective of the number of context data points, CMANPs also perform updates in constant computation and memory. In Table \ref{table:memory_complexities} and \ref{table:memory_complexities:ND_variants}, we compare the memory complexities of state-of-the-art Neural Processes, showing the efficiency gains of CMANPs.

\begin{table}[h]
\centering
\begin{tabular}{|l|ccccc|}
\hline
                     & \multicolumn{5}{c|}{\textbf{Memory Complexity}}                                                                                                                       \\ \cline{2-6}
                     & \multicolumn{1}{c|}{\textbf{Cond.}} & \multicolumn{2}{c|}{\textbf{Query}}                              & \multicolumn{2}{c|}{\textbf{Update}}                     \\ \hline
\textbf{In Terms of} & \multicolumn{1}{c|}{$|\mathcal{D}_C|$}                & \multicolumn{1}{c|}{$|\mathcal{D}_C|$} & \multicolumn{1}{c|}{$M$}              & \multicolumn{1}{c|}{$|\mathcal{D}_C|$}              & $|\mathcal{D}_U|$            \\  \hline
\textbf{TNP-D}       & \multicolumn{1}{c|}{N/A}                & \multicolumn{1}{c|}{\nocheckmark}    & \multicolumn{1}{c|}{\nocheckmark}                 & \multicolumn{1}{c|}{N/A}              & N/A              \\ 
\textbf{LBANP}       & \multicolumn{1}{c|}{\halfcheckmark}   & \multicolumn{1}{c|}{\checkmark}   & \multicolumn{1}{c|}{\halfcheckmark} & \multicolumn{1}{c|}{\halfcheckmark} & \halfcheckmark \\ \hline
\textbf{CMANP (Ours)}      & \multicolumn{1}{c|}{\checkmark}                  & \multicolumn{1}{c|}{\checkmark}   & \multicolumn{1}{c|}{\halfcheckmark}                & \multicolumn{1}{c|}{\checkmark}                & \checkmark \\  \hline
\end{tabular}
\caption{
    Comparison of Memory Complexities of top-performing Neural Processes with respect to the number of context data points $|\mathcal{D}_C|$, number of target data points in a batch $M$, and the number of new data points in an update $|\mathcal{D}_U|$.
    (Green) Checkmarks represent requiring constant memory, (Orange) half checkmarks represent requiring linear memory, and (Red) Xs represent requiring quadratic or more memory. 
    A larger table with all baselines is included in the Appendix (Table \ref{table:memory_complexity:all_baselines}).
}
\label{table:memory_complexities} 
\end{table}

\begin{table}[]
\centering
\begin{tabular}{|l|ccccc|}
\hline
                     & \multicolumn{5}{c|}{\textbf{Memory Complexity}}                                                                                                                       \\ \cline{2-6}
                     & \multicolumn{1}{c|}{\textbf{Cond.}} & \multicolumn{2}{c|}{\textbf{Query}}                              & \multicolumn{2}{c|}{\textbf{Update}}                     \\ \hline
\textbf{In Terms of} & \multicolumn{1}{c|}{$|\mathcal{D}_C|$}                & \multicolumn{1}{c|}{$|\mathcal{D}_C|$} & \multicolumn{1}{c|}{$M$}              & \multicolumn{1}{c|}{$|\mathcal{D}_C|$}              & $|\mathcal{D}_U|$            \\  \hline
\textbf{TNP-ND}      & \multicolumn{1}{c|}{N/A}                & \multicolumn{1}{c|}{\nocheckmark}    & \multicolumn{1}{c|}{\nocheckmark}                 & \multicolumn{1}{c|}{N/A}              & N/A              \\ 
\textbf{LBANP-ND}    & \multicolumn{1}{c|}{\halfcheckmark}   & \multicolumn{1}{c|}{\checkmark}   & \multicolumn{1}{c|}{\nocheckmark}                 & \multicolumn{1}{c|}{\halfcheckmark} & \halfcheckmark \\ \hline
\textbf{CMANP-AND}  & \multicolumn{1}{c|}{\checkmark}                  & \multicolumn{1}{c|}{\checkmark}   & \multicolumn{1}{c|}{\halfcheckmark}                & \multicolumn{1}{c|}{\checkmark}                & \checkmark \\ \hline
\end{tabular}
\caption{
    Comparison of Memory Complexities of Not-Diagonal extensions of Neural Processes.
}
\label{table:memory_complexities:ND_variants}
\end{table}

\subsubsection{Autoregressive Not-Diagonal Extension} \label{sec:method:CMANP_AND}

In many settings where NPs are applied such as Image Completion, the target data points are correlated and their predictive distribution are evaluated altogether. 
As such, prior works~\citep{nguyen2022transformer, feng2023latent} have proposed Not-Diagonal extensions of NPs which predicts the mean and a full covariance matrix, typically via a low-rank approximation.
This is in contrast to the vanilla (Diagonal) variants which predict the mean and a diagonal covariance matrix.
Not-Diagonal methods, however, are not scalable, requiring quadratic memory in the number of target data points due to outputting a full covariance matrix.

Leveraging the efficient updates property of CMABs, we propose CMANP-AND (Autoregressive Not-Diagonal).
During training, CMANP-AND follows the framework of prior Not-Diagonal variants. 
When deployed, the model is treated as an autoregressive model that makes predictions in blocks of size $b_Q$ data points. 
For each block prediction, a mean and full covariance matrix is computed via a low-rank approximation.  
Sampled predictions of prior blocks are used to make predictions for later blocks.
The first block is sampled as follows: $\hat{Y}_1 \sim \mathcal{N} (\mu_\theta(\mathcal{D}_C^{0}, X_1), \Sigma_\theta(\mathcal{D}_C^{0}, X_1))$ where $\hat{Y}_1 = \hat{y}_{N+1: N+b_Q}$ and $X_1 = x_{N + 1: N + b_Q}$. Afterwards, by leveraging the efficient update mechanism, CMANP-AND performs an update using the sampled predictions $\{(x_i, \hat{y}_i)\}^{N+b_Q}_{N+1}$ as new context data points, meaning that CMANP-AND is now conditioned on a new context dataset $\mathcal{D}_C^{1}$ where $\mathcal{D}_C^{1} = \mathcal{D}_C^{0} \cup \{(x_i, \hat{y}_i)\}^{N+b_Q}_{N+1}$.
The general formulation is as follows: $\hat{Y}_{k+1} \sim \mathcal{N} (\mu_\theta(\mathcal{D}_C^k, X_{k+1}), \Sigma_\theta(\mathcal{D}_C^k, X_{k+1}))$ 
where $k$ is the number of blocks already processed, $\hat{Y}_{k+1} = \hat{y}_{N + k b_Q  + 1: N + (k +1) b_Q}$, $X_{k+1} = x_{N + k b_Q + 1: N + (k+1) b_Q}$, and $\mathcal{D}_C^k = \{(x_i, y_i)\}_{i=1}^N \cup \{(x_i, \hat{y}_i)\}^{N+k b_Q}_{N+1}$ is the context dataset.
The hyperparameter $b_Q$ controls (1) the computational cost of the model in terms of memory and sequential computation length and (2) the performance of the model. 
Lower values of $b_Q$ allow for modelling more complex distributions, offering better performance but requiring more forward passes of the model. 
Since $b_Q$ is a constant, this Autoregressive Not-Diagonal extension makes predictions significantly more efficiently than the prior Not-Diagonal variants which were quadratic in memory. As such, CMANP-AND can scale to a larger number of data points than prior methods (LBANP-ND and TNP-ND). Big-$\mathcal{O}$ complexity analysis is included in Appendix \ref{appendix:proof:complexity_analysis_cmanp_and}.

\section{Experiments} \label{sec:experiments}

In this section, we aim to evaluate the empirical performance of CMANPs and provide an analysis of CMANPs. 
To do so, we compare CMANPs against a large variety of members of the Neural Process family on standard NP benchmarks: image completion and meta-regression. 
Specifically, we compare against Conditional Neural Processes (CNPs)~\citep{garnelo2018conditional}, Neural Processes (NPs)~\citep{garnelo2018neural}, Bootstrapping Neural Processes (BNPs)~\citep{lee2020bootstrapping}, (Conditional) Attentive Neural Processes (C)ANPs~\citep{kim2019attentive}, Bootstrapping Attentive Neural Processes (BANPs)~\citep{lee2020bootstrapping}, Latent Bottlenecked Attentive Neural Processes (LBANPs)~\citep{feng2023latent}, and Transformer Neural Processes (TNPs)~\citep{nguyen2022transformer}.
We also compare against the Not-Diagonal extensions of the state-of-the-art methods (i.e., LBANP-ND and TNP-ND).

For consistency, we set the number of latents (i.e., bottleneck size) $|L_I| = |L_B| = 128$ across all experiments. We also set $b_Q = 5$. To fairly compare iterative attention and CMABs, we report results for LBANPs with the same sized bottleneck (i.e., number of latents $L = 128$) as CMANPs across all experiments.
Due to space limitations, additional hyperparameters and experiments are included in the appendix\footnote{Code: \href{https://github.com/BorealisAI/constant-memory-anp}{https://github.com/BorealisAI/constant-memory-anp}.}.

\subsection{Image Completion}

In these experiments, we consider the image completion setting.
The model is given a set of pixel values of an image and aims to predict the remaining pixels of the image. Each image corresponds to a unique function~\citep{garnelo2018neural}. In these experiments, the $x$ values are rescaled to [-1, 1] and the $y$ values are rescaled to $[-0.5, 0.5]$. For each task, a randomly selected set of pixels are selected as context data points and target data points. 

EMNIST~\citep{cohen2017emnist} comprises black and white images of handwritten letters of $32 \times 32$ resolution. $10$ classes are used for training. The context and target data points are sampled according to $N \sim \mathcal{U}[3, 197)$ and $M \sim \mathcal{U}[3, 200-N)$  respectively.
CelebA~\citep{liu2015faceattributes} comprises coloured images of celebrity faces. Methods are evaluated on various resolutions to show the scalability of the methods. 
In CelebA32, images are downsampled to $32 \times 32$ and the number of context and target data points are sampled according to $N \sim \mathcal{U}[3, 197)$ and $M \sim \mathcal{U}[3, 200-N)$ respectively. 
In CelebA64, the images are down-sampled to $64 \times 64$  and $N \sim \mathcal{U}[3, 797)$ and $M \sim \mathcal{U}[3, 800-N)$.
In CelebA128, the images are down-sampled to $128 \times 128$ and $N \sim \mathcal{U}[3, 1597)$ and $M \sim \mathcal{U}[3, 1600-N)$.

\begin{table*}[ht]
\centering
\begin{tabular}{|c|ccc|cc|}
\hline
\multirow{2}{*}{Method}      & \multicolumn{3}{c|}{CelebA}                & \multicolumn{2}{c|}{EMNIST}  \\
                             & 32x32        & 64x64        & 128x128      & Seen (0-9)  & Unseen (10-46) \\ \hline
CNP                          & 2.15 ± 0.01  & 2.43 ± 0.00  & 2.55 ± 0.02  & 0.73 ± 0.00 & 0.49 ± 0.01    \\
CANP                         & 2.66 ± 0.01  & 3.15 ± 0.00  & ---          & 0.94 ± 0.01 & 0.82 ± 0.01    \\
NP                           & 2.48 ± 0.02  & 2.60 ± 0.01  & 2.67 ± 0.01  & 0.79 ± 0.01 & 0.59 ± 0.01    \\
ANP                          & 2.90 ± 0.00  & ---          & ---          & 0.98 ± 0.00 & 0.89 ± 0.00    \\
BNP                          & 2.76 ± 0.01  & 2.97 ± 0.00  & ---          & 0.88 ± 0.01 & 0.73 ± 0.01    \\
BANP                         & 3.09 ± 0.00  & ---          & ---          & 1.01 ± 0.00 & 0.94 ± 0.00    \\
TNP-D                        & 3.89 ± 0.01  & 5.41 ± 0.01  & ---          & \textbf{1.46 ± 0.01} & \textbf{1.31 ± 0.00}    \\
LBANP                & 3.97 ± 0.02  & 5.09 ± 0.02  & 5.84 ± 0.01  & 1.39 ± 0.01 & 1.17 ± 0.01    \\ \hline
CMANP (Ours)     & 3.93 ± 0.05 & 5.02 ± 0.14 & 5.55 ± 0.01 & 1.36 ± 0.01         & 1.09 ± 0.01            \\ \hline
TNP-ND                       & 5.48 ± 0.02  & ---          & ---          & \textbf{1.50 ± 0.00} & \textbf{1.31 ± 0.00}    \\
LBANP-ND               & 5.57 ± 0.03  & ---          & ---          & 1.42 ± 0.01 & 1.14 ± 0.01    \\ \hline
CMANP-AND (Ours) & \textbf{6.31 ± 0.04} & \textbf{6.96 ± 0.07} & \textbf{7.15 ± 0.14} & \textbf{1.48 ± 0.03} & 1.19 ± 0.03    \\ \hline
\end{tabular}
\caption{Image Completion Experiments. Each method is evaluated with 5 different seeds according to the log-likelihood (higher is better). The "dash" represents methods that could not be run because of the large memory requirement.}
\label{table:image_completion}
\end{table*}

\textbf{Results.} 
All NP baselines (see Table \ref{table:image_completion}) were able to be evaluated on CelebA (32 x 32) and EMNIST. 
However, many baselines were not able to be trained on CelebA (64 x 64) and CelebA (128 x 128) in the available GPU memory due to requiring a quadratic amount of memory including all prior Not-Diagonal extensions of NPs.
In contrast, CMANP(-AND) was not affected by this limitation due to being significantly more memory efficient than prior Not-Diagonal variants.
The results show that CMANP-AND achieves clear state-of-the-art results on CelebA (32x32), CelebA (64x64), and CelebA (128x128) while being scalable to more data points than prior Not-Diagonal variants.
Furthermore, CMANP-AND achieves results competitive with state-of-the-art EMNIST.

\begin{figure*}[h]
     \centering
     \includegraphics[width=0.31\textwidth]{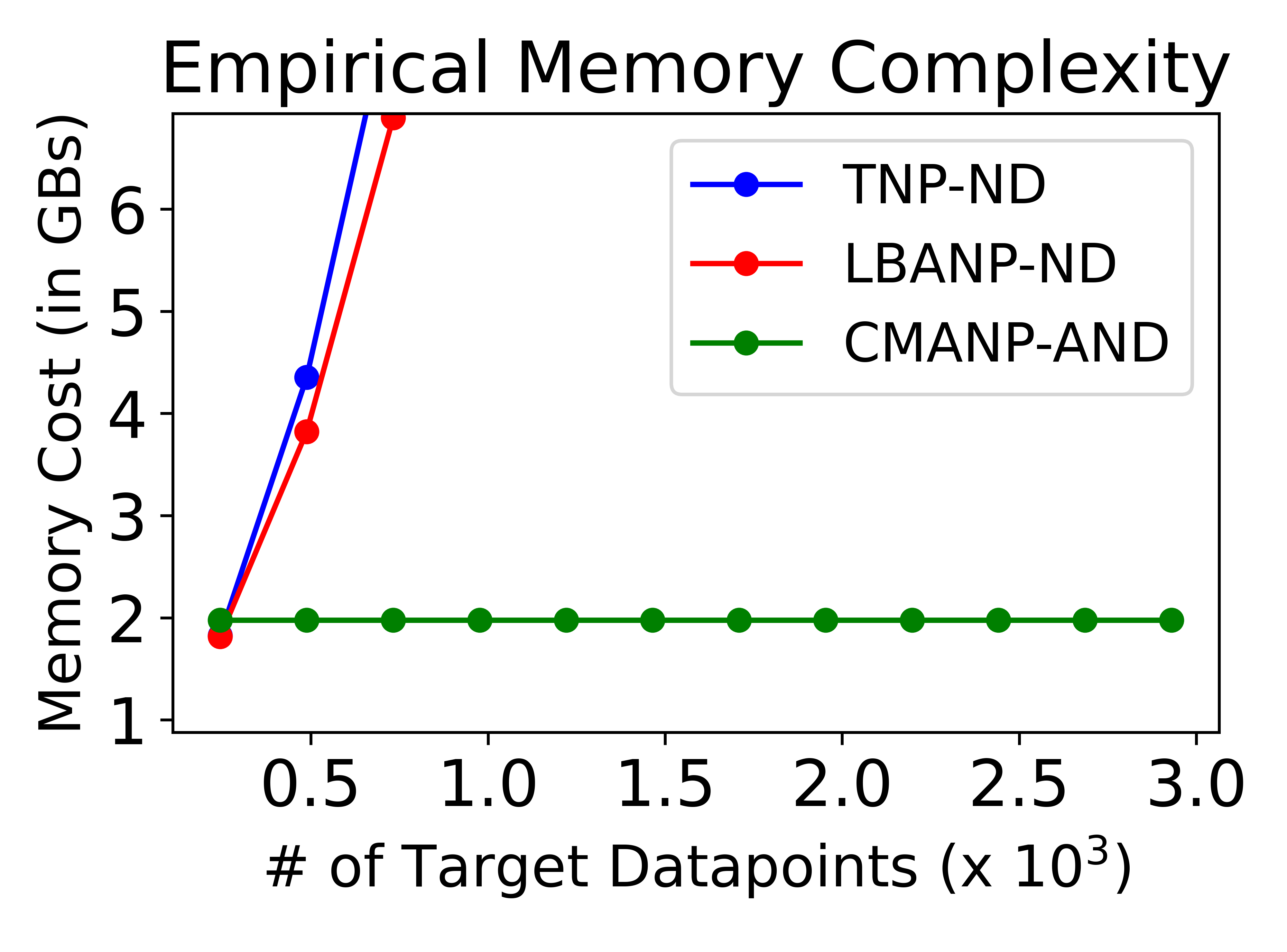}
     \includegraphics[width=0.31\textwidth]
     {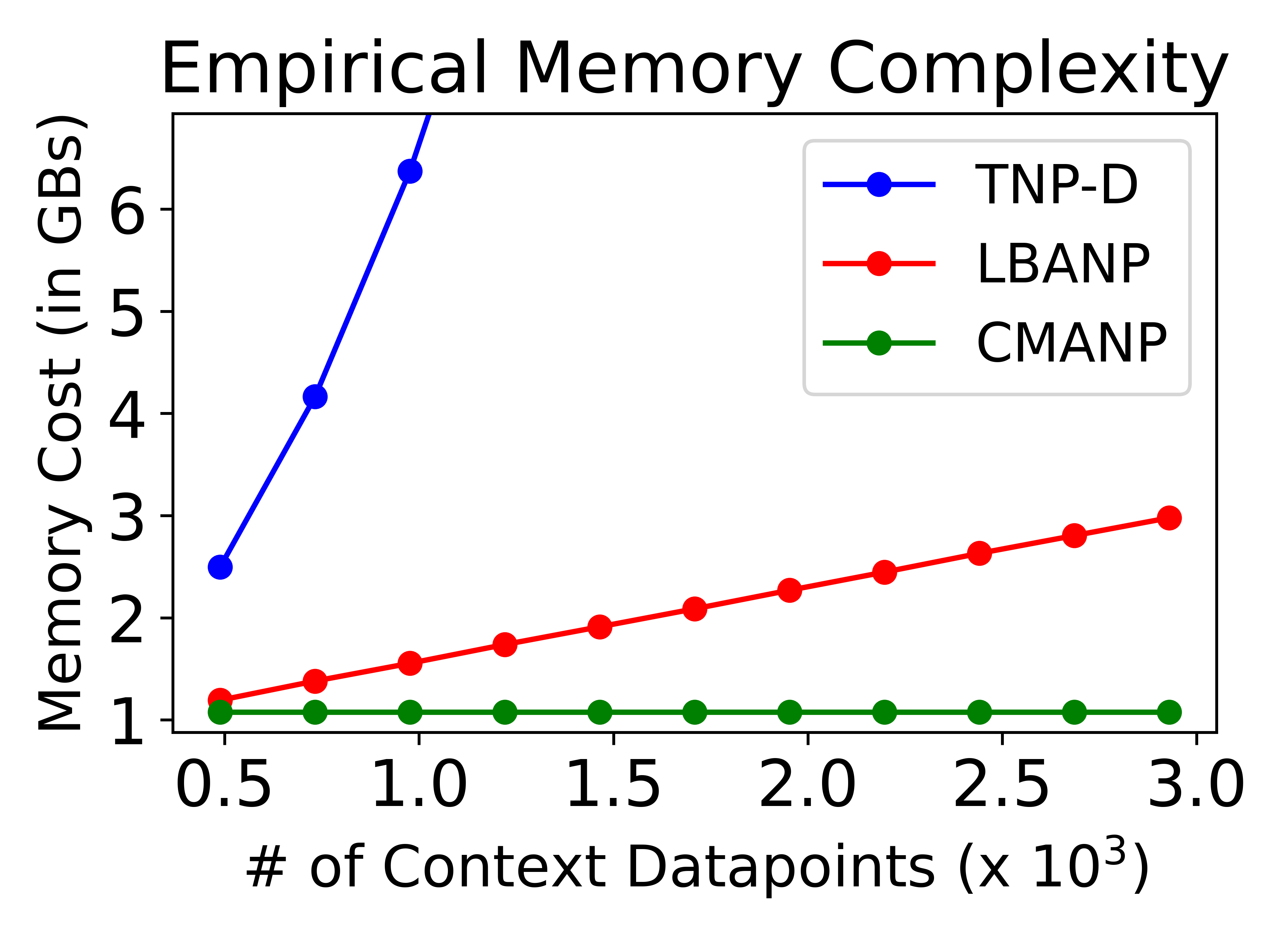}
     \includegraphics[width=0.31\textwidth]{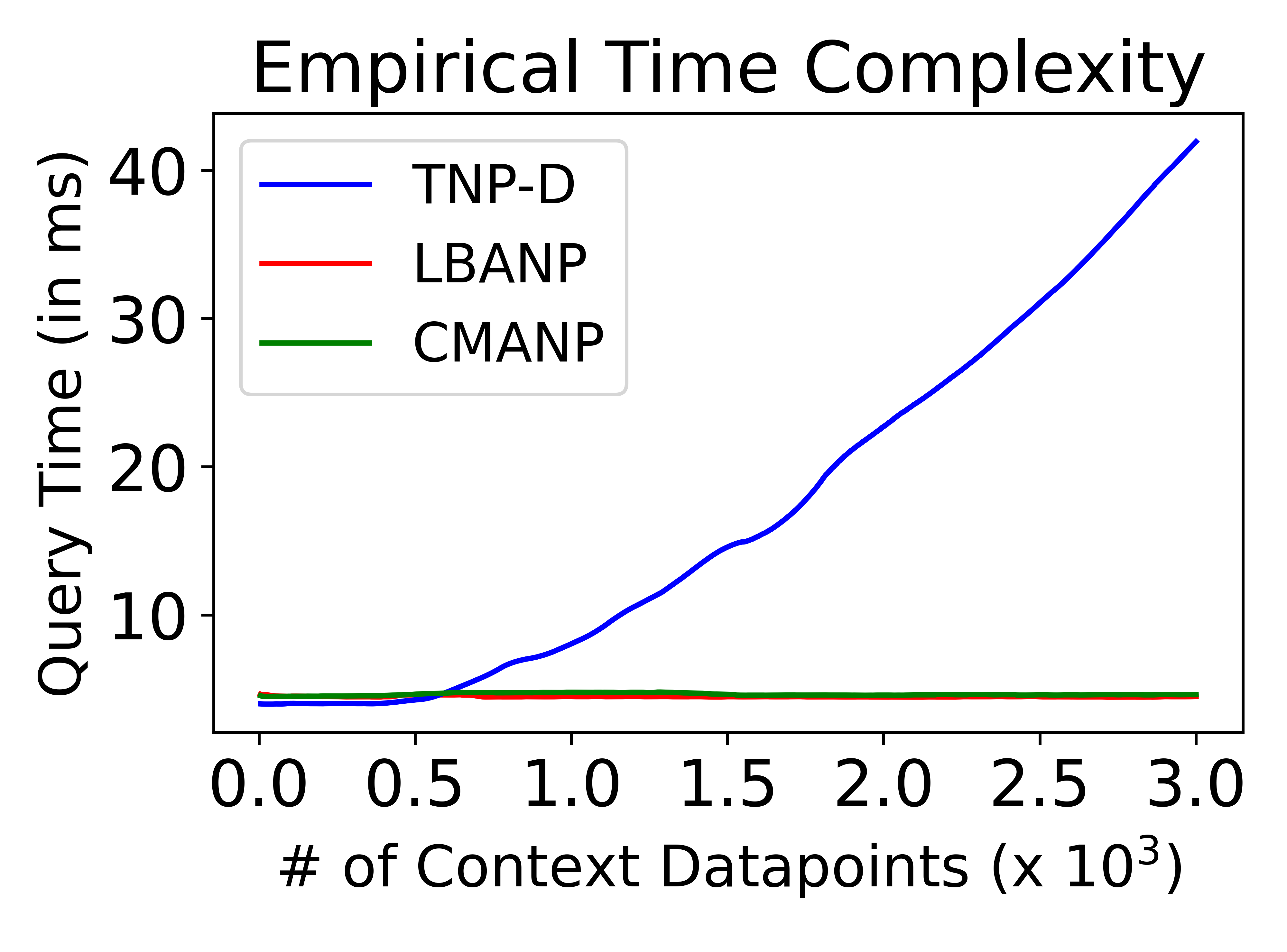}
    \caption{ Computational Complexity Comparison Plots. (Left)  Empirical memory usage comparison of CMANP-AND with state-of-the-art Not-Diagonal NPs. (Middle) Empirical memory usage comparison of CMANPs with state-of-the-art NPs.
    (Right) Empirical runtime comparison of CMANPs with state-of-the-art NPs.
    }
    \label{fig:analyses:computational_efficiency}
\end{figure*}

\textbf{Empirical Computational Complexity Results.} In Figure \ref{fig:analyses:computational_efficiency}, we analyze the empirical memory and runtime used by CMANPs compared with various state-of-the-art Neural Process models. Comparing Not-Diagonal extensions of NPs (Figure \ref{fig:analyses:computational_efficiency} (Left)), we see that the memory usage of both TNP-ND and LBANP-ND scale quadratically with respect to the number of target data points, making these methods infeasible to be applied to settings with a larger number of target data points as shown in our CelebA experiments. In contrast, CMANP-AND can scale to a far larger number of target data points. Comparing the vanilla variants of NPs (Figure \ref{fig:analyses:computational_efficiency} (Middle)), we see that TNP-D (transformer-based model) scales quadratically with respect to the number of context data points while LBANP (Perceiver-based model) scales linearly. In contrast, CMANP (CMAB-based model) only requires a low constant amount of memory regardless of the number of context data points. 
As a result, we can see that CMANPs use a significantly less amount of memory, allowing them to scale to more data points while achieving results competitive with state-of-the-art. 

In terms of runtime (Figure \ref{fig:analyses:computational_efficiency} (Right)), LBANPs and CMANPs are comparable as they both only require retrieving information from a fixed-size set of latents. In contrast, TNPs are quadratic due to leveraging transformers during queries. 
Due to space limitations, we include in the appendix the runtime complexity plot of the update operation which shows that CMANPs' efficient update mechanism is significantly more efficient than that of prior state-of-the-art.

\subsection{Meta-Regression}

In this experiment, the goal is to model an unknown function $f$ and make predictions for a batch of $M$ target data points given a batch of $N$ context data points. 
During each training epoch, a batch of $B=16$ functions are sampled from a GP prior with an RBF kernel $f_i \sim GP(m, k)$ where $m(x) = 0$ and $k(x, x')= \sigma_f^2 \exp(\frac{-(x-x')^2}{2l^2})$. The hyperparameters are sampled according to  $l \sim \mathcal{U}[0.6, 1.0)$, $\sigma_f \sim \mathcal{U}[0.1, 1.0)$, $N \sim \mathcal{U}[3, 47)$, and $M \sim \mathcal{U}[3, 50-N)$.
After training, the models are evaluated according to the log-likelihood of the targets on functions sampled from GPs with RBF and Matern $5/2$ kernels. 

\textbf{Results.} As shown in Table \ref{table:1d_regression}, CMANP-AND achieves comparable results to TNP-ND and outperforms all other baselines by a significant margin while only requiring constant memory.  Furthermore, we see that the vanilla version of CMANP (CMAB-based model) and LBANP (Perceiver-based model~\citep{jaegle2021perceiver}) achieve similar performance, showing that CMANPs achieve competitive performance while being significantly more efficient (constant memory and constant computation updates). 

\begin{table}
\centering
\begin{tabular}{|c|cc|}
\hline
Method                       & RBF          & Matern 5/2   \\ \hline
CNP                          & 0.26 ± 0.02  & 0.04 ± 0.02  \\
CANP                         & 0.79 ± 0.00  & 0.62 ± 0.00  \\
NP                           & 0.27 ± 0.01  & 0.07 ± 0.01  \\
ANP                          & 0.81 ± 0.00  & 0.63 ± 0.00  \\
BNP                          & 0.38 ± 0.02  & 0.18 ± 0.02  \\
BANP                         & 0.82 ± 0.01  & 0.66 ± 0.00  \\
TNP-D                        & 1.39 ± 0.00  & 0.95 ± 0.01  \\
LBANP                & 1.27 ± 0.02  & 0.85 ± 0.02  \\ \hline
CMANP (Ours)      & 1.24 ± 0.01 & 0.80 ± 0.01 \\ \hline
TNP-ND                       & \textbf{1.46 ± 0.00}  & \textbf{1.02 ± 0.00}  \\
LBANP-ND             & 1.24 ± 0.03          & 0.78 ± 0.02         \\ \hline
CMANP-AND (Ours)   & \textbf{1.48 ± 0.03} & 0.96 ± 0.01 \\ \hline
\end{tabular}
\caption{1-D Meta-Regression Experiments with log-likelihood metric (higher is better). 
}
\vspace{-5mm}
\label{table:1d_regression}
\end{table}

\subsubsection{Analysis}
\label{sec:analysis}

\begin{figure*}[h]
     \centering
     \includegraphics[width=0.31\textwidth]{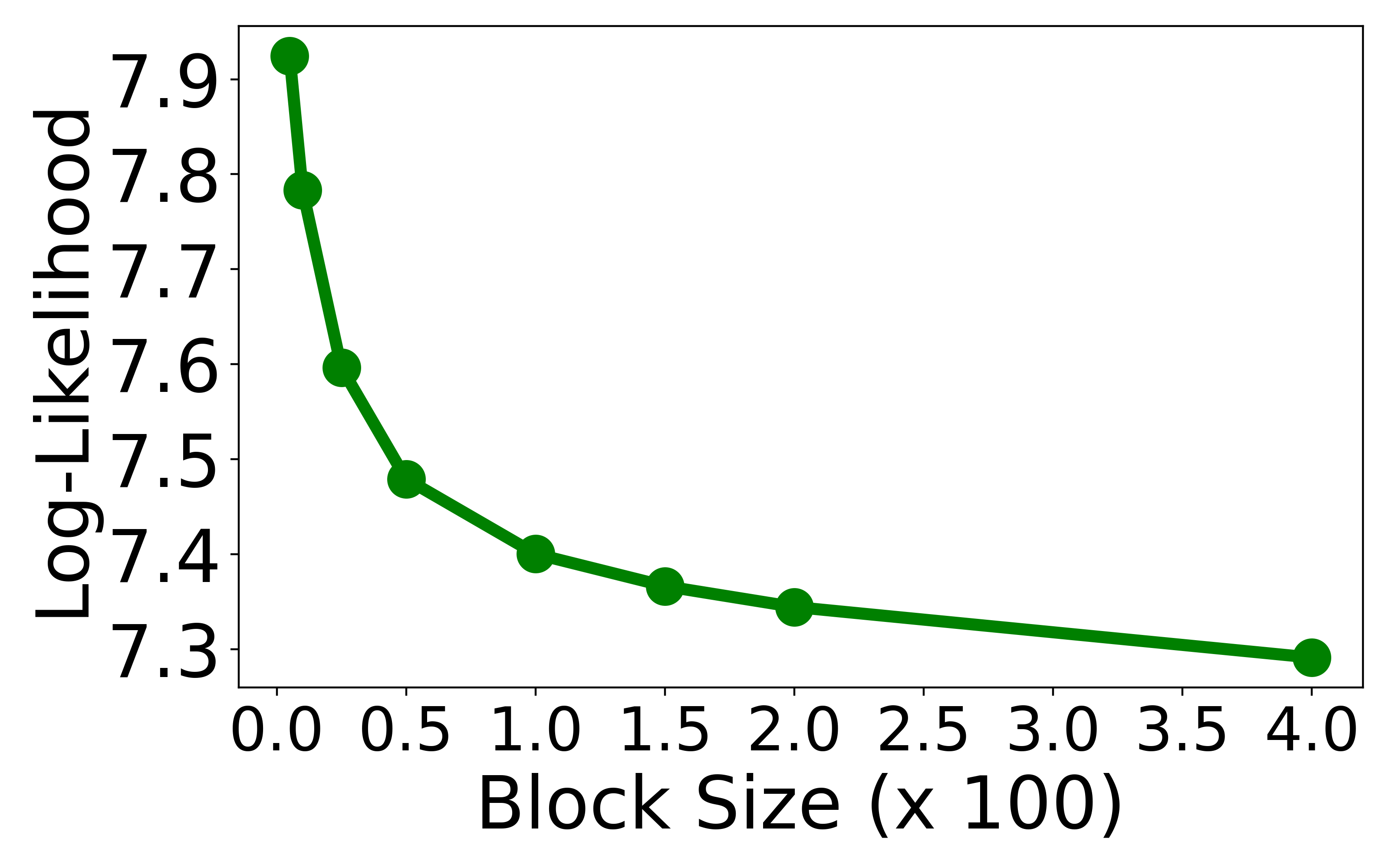}
     \includegraphics[width=0.31\textwidth]
     {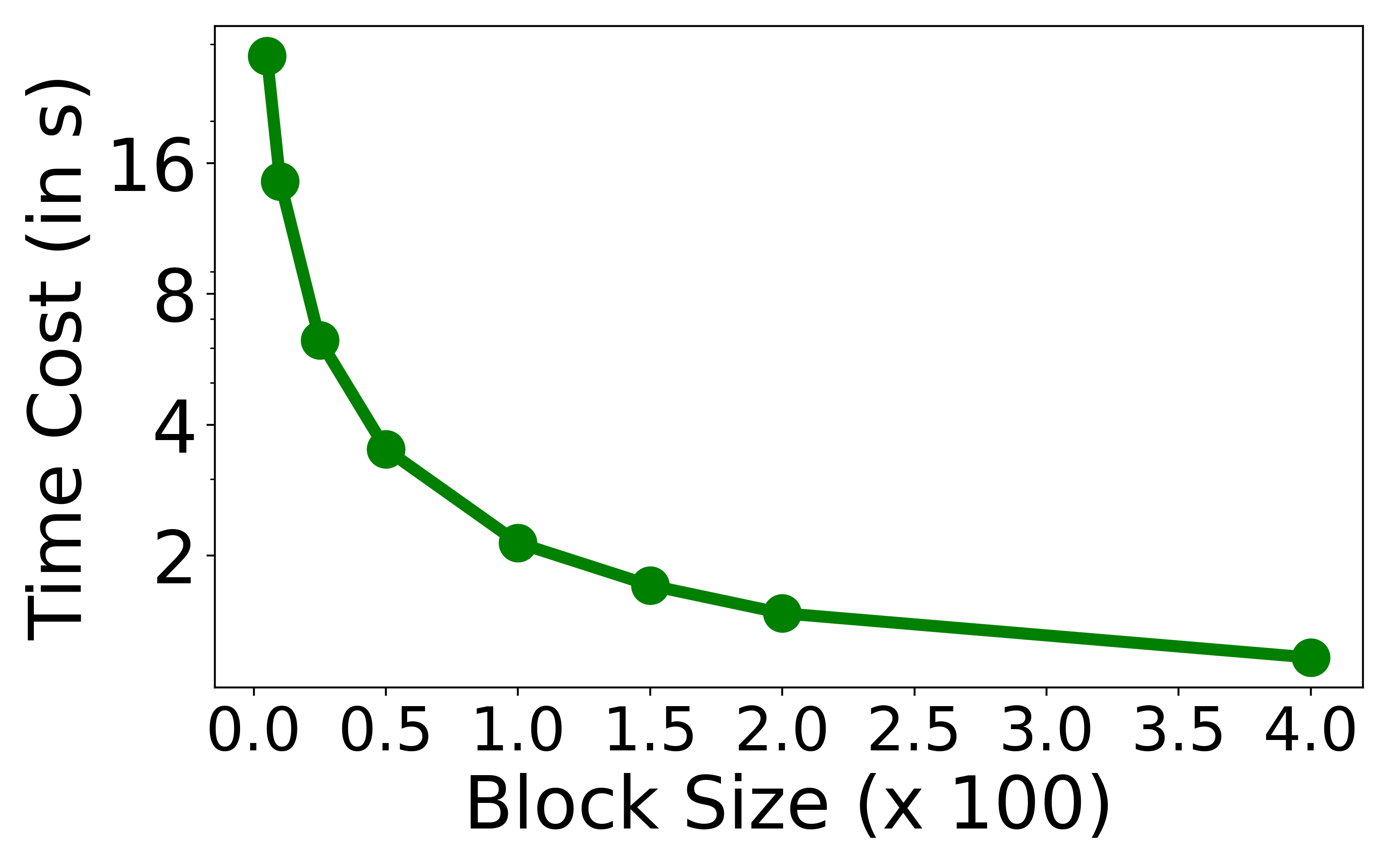}
     \includegraphics[width=0.31\textwidth]{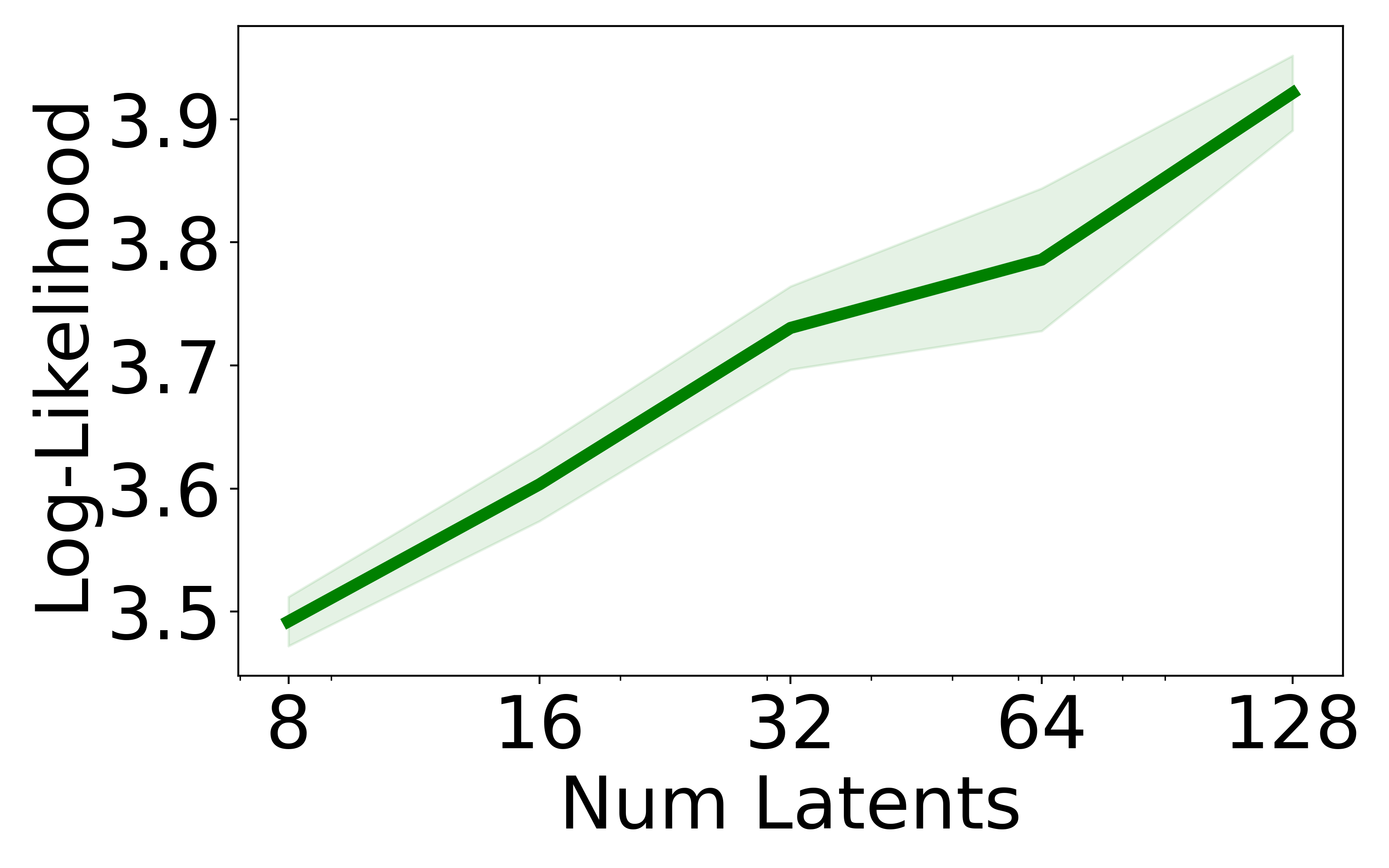}
    \caption{ Analyses Plots. (Left)  CMANP's runtime relative to the predictive block size ($b_Q$). (Middle) CMANP's performance relative to the predictive block size ($b_Q$).
    (Right) CMANP's performance relative to the number of latent vectors ($|L_I| = |L_B|$).
    }
    \label{fig:analyses:hyperparameter_impact}
\end{figure*}

\textbf{Effect of Block Size ($b_Q$):} In Figure \ref{fig:analyses:hyperparameter_impact} (Left and Middle), we evaluate the test-time performance and runtime for CMANP-AND with respect to the block size ($b_Q$). 
The smaller the block size, the better the performance. This is expected as the autoregressive nature of CMANP-AND allows for more flexible predictive distribution and hence better performance. 
This, however, comes at the cost of an increased time complexity.
In conjunction, these plots show that there is a trade-off between the time cost and performance. 
As such, $b_Q$ is a hyperparameter that can be adjusted based on the available resources.

\textbf{Varying Number of Latents:}
Figure \ref{fig:analyses:hyperparameter_impact} (Right) evaluates the result of varying the number of latents ($|L_I| = |L_B|$), showing that increasing the number of latents (i.e., the size of the bottleneck) improves the performance of the model.

\textbf{Extending CMAB beyond Neural Processes:}
Constant Memory Attention Blocks (1) do not leverage any modality-specific components and (2) are permutation invariant by default like Transformers. As such, CMABs can naturally extend beyond that of Neural Processes. 
As a proof of concept, we experiment with extending CMABs to next-event prediction (Temporal Point Processes (TPPs)) by replacing the transformer in Transformer Hawkes Process (THP)~\citep{zuo2020transformer}, a popular method in the TPP literature, with our proposed Constant Memory Attention Block, resulting in the Constant Memory Hawkes Process. In Table \ref{table:tpp}, we see that CMHP achieves comparable results with that of THP on a popular dataset in the TPP literature: Reddit.  Crucially, CMHP only requires constant memory unlike the quadratic memory required by that of THP.
Furthermore, CMHP efficiently updates its model with new data as it arrives over time which is typical in event sequence data, making it significantly more efficient than the quadratic computation required by THP.

\begin{table}[ht]
\resizebox{\linewidth}{!}{%
\centering
\begin{tabular}{|l|ccc|}
\hline
\multicolumn{1}{|c|}{\multirow{2}{*}{Model}} & \multicolumn{3}{c|}{Reddit}                                                                                       \\ \cline{2-4} 
\multicolumn{1}{|c|}{} & \multicolumn{1}{c|}{RMSE}          & \multicolumn{1}{c|}{NLL}           & ACC                                     \\ \hline
THP                    & \multicolumn{1}{c|}{\textbf{0.238 ± 0.028}} & \multicolumn{1}{c|}{\textbf{0.268 ± 0.098}} & \textbf{0.610 ± 0.002} \\ \hline
CMHP            & \multicolumn{1}{c|}{\textbf{0.262 ± 0.037}} & \multicolumn{1}{c|}{\textbf{0.528 ± 0.209}} & \textbf{0.609 ± 0.003} \\ \hline
\end{tabular}
}
\caption{Temporal Point Processes Experiments.}
\label{table:tpp}
\end{table}

\section{Related Work} \label{sec:related_work}

Meta-learning has two learning phases (inner/task and outer/meta). In this work, we consider Neural Processes (NPs), a member of the model-based (or black-box) meta-learning family. Model-based/black-box methods' perform their inner learning phase via a forward pass through a model, often a neural network. Several architectures have been proposed for the inner learning phase such as convolutional networks~\citep{mishra2018simple}, recurrent networks~\citep{zintgraf2021varibad}, attention~\citep{feng2023latent}, and hypernetworks~\citep{chen2022transformers}. For an in-depth overview of the meta-learning literature, we refer the reader to the following meta-learning survey~\citep{hospedales2022meta}.

NPs have been applied to a wide range of applications which include sequence data~\citep{singh2019sequential, willi2019recurrent}, modelling stochastic physics fields~\citep{holderrieth2021equivariant}, robotics~\citep{chen2022meta, li2022category}, event prediction~\citep{bae2023meta}, and climate modeling~\citep{vaughan2021convolutional}.
In doing so, there have been several methods proposed for encoding the context dataset. For example, CNPs~\citep{garnelo2018conditional} encode the context set via a deep sets encoder~\citep{zaheer2017deep}, 
NPs~\citep{garnelo2018neural} propose to encode functional stochasticity via a latent variable. 
ConvCNPs~\citep{gordon2019convolutional} use convolutions to build in translational equivariance. 
ANPs~\citep{kim2019attentive}, LBANPs~\citep{feng2023latent}, and TNPs~\citep{nguyen2022transformer} use various kinds of attention. 
Recent work~\citep{bruinsma2023autoregressive} builds on CNPs and ConvCNPs by proposing to make them autoregressive at deployment.
For an in-depth overview of NPs, we refer the reader to the recent survey work~\citep{jha2022neural}.

Transformers~\citep{vaswani2017attention} have achieved a large amount of success in a wide range of applications. 
However, the quadratic scaling of Transformers limits their applications. As such, there have been many follow-up works on efficient variants. However, very few works have achieved constant memory complexity.
\citet{rabe2022self} showed that self-attention can be computed in constant memory at the expense of an overall quadratic computation. 
In contrast, CMABs only require linear computation and constant memory, making it significantly more efficient.
\citet{wu2022memformer} and \citet{peng2023rwkv} proposed Memformer and RWKV respectively, constant memory versions of transformer for sequence modelling problems. However, these methods are dependent on the order of the tokens, limiting their applications.
In contrast, CMABs are by default permutation-invariant, allowing them to be more flexible in applications with their data modality.
For an in-depth overview of follow-up works to Transformers, we refer the reader to the recent survey works~\citep{khan2022transformers,lin2022survey}.

Although CMABs have an efficient update mechanism reminiscent of RNNs~\citep{cho2014properties, chung2014empirical, hochreiter1997long}, their applications are different. RNNs are sensitive to input order, making their ideal setting applications which use sequential data. In contrast, by design, CMABs are by default permutation-invariant. Due to their long computation graph, RNNs also have issues such as vanishing gradients, making training these models with a large number of data points difficult. CMABs do not have issues with vanishing gradients since their ability to update efficiently is a fixed property of the module rather than RNN's learned mechanism.

\section{Conclusion} \label{sec:conclusion}

In this work, we showed that Cross Attention can be updated efficiently with new context data and computed in a constant amount of memory irrespective of the number of context data points. 
Leveraging these properties, we introduced CMAB (Constant Memory Attention Block), a novel attention block that (1) is permutation invariant, (2) computes its output in constant memory, and (3) performs updates in constant computation. 
Building on CMAB, we proposed Constant Memory Attentive Neural Processes (CMANPs), a new memory-efficient NP variant.
Leveraging the efficient updates property of CMAB, we introduced CMANP-AND (Autoregressive Not-Diagonal extension).
Empirically, we show that CMANP(-AND) achieves state-of-the-art results while being significantly more efficient than prior state-of-the-art methods. In our analysis, we showed that increasing the size of the latent bottleneck can improve CMANPs' performance. 
Lastly, we showed a proof of concept extending CMABs beyond Neural Processes.

\section*{Impact Statement} \label{sec:broader_impact}

In this paper, we improve the memory efficiency of Neural Processes, a family of deep learning models for uncertainty estimation.
In order to deploy deep learning models to real-world settings, it is imperative that the models be memory efficient whilst simultaneously being able to capture their predictive uncertainty; incorrect or overconfident predictions can have dire consequences in mission- or safety-critical settings~\citep{tambon2022certify}.
Furthermore, designing memory efficient deep learning models is important for a variety of reasons, for example:
(1) With the growing popularity of low-memory/compute domains (e.g., IoT devices and mobile phones), it is crucial that deep learning models be designed memory efficiently in order to be deployable on the device. (2) In addition, memory access is energy intensive~\citep{zhi2022memory}. Since many safety-critical settings (e.g., autonomous vehicles and mobile robots in disaster response) have a limited energy supply, it is therefore crucial to develop memory efficient models that can simultaneously model predictive uncertainty such as Neural Processes.

\bibliography{icml}

\begin{thebibliography}{44}
\providecommand{\natexlab}[1]{#1}
\providecommand{\url}[1]{\texttt{#1}}
\expandafter\ifx\csname urlstyle\endcsname\relax
  \providecommand{\doi}[1]{doi: #1}\else
  \providecommand{\doi}{doi: \begingroup \urlstyle{rm}\Url}\fi

\bibitem[Bae et~al.(2023)Bae, Ahmed, Tung, and Oliveira]{bae2023meta}
Bae, W., Ahmed, M.~O., Tung, F., and Oliveira, G.~L.
\newblock Meta temporal point processes.
\newblock In \emph{International Conference on Learning Representations}, 2023.
\newblock URL \url{https://openreview.net/forum?id=QZfdDpTX1uM}.

\bibitem[Bruinsma et~al.(2023)Bruinsma, Markou, Requeima, Foong, Andersson, Vaughan, Buonomo, Hosking, and Turner]{bruinsma2023autoregressive}
Bruinsma, W., Markou, S., Requeima, J., Foong, A. Y.~K., Andersson, T., Vaughan, A., Buonomo, A., Hosking, S., and Turner, R.~E.
\newblock Autoregressive conditional neural processes.
\newblock In \emph{The Eleventh International Conference on Learning Representations}, 2023.
\newblock URL \url{https://openreview.net/forum?id=OAsXFPBfTBh}.

\bibitem[Chen et~al.(2022)Chen, Gao, Vien, Ziesche, and Neumann]{chen2022meta}
Chen, R., Gao, N., Vien, N.~A., Ziesche, H., and Neumann, G.
\newblock Meta-learning regrasping strategies for physical-agnostic objects.
\newblock \emph{arXiv preprint arXiv:2205.11110}, 2022.

\bibitem[Chen \& Wang(2022)Chen and Wang]{chen2022transformers}
Chen, Y. and Wang, X.
\newblock Transformers as meta-learners for implicit neural representations.
\newblock In \emph{European Conference on Computer Vision}, pp.\  170--187. Springer, 2022.

\bibitem[Cho et~al.(2014)Cho, Van~Merri{\"e}nboer, Bahdanau, and Bengio]{cho2014properties}
Cho, K., Van~Merri{\"e}nboer, B., Bahdanau, D., and Bengio, Y.
\newblock On the properties of neural machine translation: Encoder-decoder approaches.
\newblock \emph{arXiv preprint arXiv:1409.1259}, 2014.

\bibitem[Chung et~al.(2014)Chung, Gulcehre, Cho, and Bengio]{chung2014empirical}
Chung, J., Gulcehre, C., Cho, K., and Bengio, Y.
\newblock Empirical evaluation of gated recurrent neural networks on sequence modeling.
\newblock \emph{arXiv preprint arXiv:1412.3555}, 2014.

\bibitem[Cohen et~al.(2017)Cohen, Afshar, Tapson, and Van~Schaik]{cohen2017emnist}
Cohen, G., Afshar, S., Tapson, J., and Van~Schaik, A.
\newblock Emnist: Extending mnist to handwritten letters.
\newblock In \emph{2017 international joint conference on neural networks (IJCNN)}, pp.\  2921--2926. IEEE, 2017.

\bibitem[Feng et~al.(2023)Feng, Hajimirsadeghi, Bengio, and Ahmed]{feng2023latent}
Feng, L., Hajimirsadeghi, H., Bengio, Y., and Ahmed, M.~O.
\newblock Latent bottlenecked attentive neural processes.
\newblock In \emph{International Conference on Learning Representations}, 2023.
\newblock URL \url{https://openreview.net/forum?id=yIxtevizEA}.

\bibitem[Garnelo et~al.(2018{\natexlab{a}})Garnelo, Rosenbaum, Maddison, Ramalho, Saxton, Shanahan, Teh, Rezende, and Eslami]{garnelo2018conditional}
Garnelo, M., Rosenbaum, D., Maddison, C., Ramalho, T., Saxton, D., Shanahan, M., Teh, Y.~W., Rezende, D., and Eslami, S.~A.
\newblock Conditional neural processes.
\newblock In \emph{International Conference on Machine Learning}, pp.\  1704--1713. PMLR, 2018{\natexlab{a}}.

\bibitem[Garnelo et~al.(2018{\natexlab{b}})Garnelo, Schwarz, Rosenbaum, Viola, Rezende, Eslami, and Teh]{garnelo2018neural}
Garnelo, M., Schwarz, J., Rosenbaum, D., Viola, F., Rezende, D.~J., Eslami, S., and Teh, Y.~W.
\newblock Neural processes.
\newblock \emph{arXiv preprint arXiv:1807.01622}, 2018{\natexlab{b}}.

\bibitem[Gordon et~al.(2019)Gordon, Bruinsma, Foong, Requeima, Dubois, and Turner]{gordon2019convolutional}
Gordon, J., Bruinsma, W.~P., Foong, A.~Y., Requeima, J., Dubois, Y., and Turner, R.~E.
\newblock Convolutional conditional neural processes.
\newblock In \emph{International Conference on Learning Representations}, 2019.

\bibitem[Hochreiter \& Schmidhuber(1997)Hochreiter and Schmidhuber]{hochreiter1997long}
Hochreiter, S. and Schmidhuber, J.
\newblock Long short-term memory.
\newblock \emph{Neural computation}, 9:\penalty0 1735--80, 12 1997.
\newblock \doi{10.1162/neco.1997.9.8.1735}.

\bibitem[Holderrieth et~al.(2021)Holderrieth, Hutchinson, and Teh]{holderrieth2021equivariant}
Holderrieth, P., Hutchinson, M.~J., and Teh, Y.~W.
\newblock Equivariant learning of stochastic fields: Gaussian processes and steerable conditional neural processes.
\newblock In \emph{International Conference on Machine Learning}, pp.\  4297--4307. PMLR, 2021.

\bibitem[Hospedales et~al.(2022)Hospedales, Antoniou, Micaelli, and Storkey]{hospedales2022meta}
Hospedales, T., Antoniou, A., Micaelli, P., and Storkey, A.
\newblock Meta-learning in neural networks: A survey.
\newblock \emph{IEEE transactions on pattern analysis and machine intelligence}, 44\penalty0 (9):\penalty0 5149--5169, 2022.

\bibitem[Jaegle et~al.(2021{\natexlab{a}})Jaegle, Borgeaud, Alayrac, Doersch, Ionescu, Ding, Koppula, Zoran, Brock, Shelhamer, et~al.]{jaegle2021perceiverio}
Jaegle, A., Borgeaud, S., Alayrac, J.-B., Doersch, C., Ionescu, C., Ding, D., Koppula, S., Zoran, D., Brock, A., Shelhamer, E., et~al.
\newblock Perceiver io: A general architecture for structured inputs \& outputs.
\newblock In \emph{International Conference on Learning Representations}, 2021{\natexlab{a}}.

\bibitem[Jaegle et~al.(2021{\natexlab{b}})Jaegle, Gimeno, Brock, Vinyals, Zisserman, and Carreira]{jaegle2021perceiver}
Jaegle, A., Gimeno, F., Brock, A., Vinyals, O., Zisserman, A., and Carreira, J.
\newblock Perceiver: General perception with iterative attention.
\newblock In \emph{International conference on machine learning}, pp.\  4651--4664. PMLR, 2021{\natexlab{b}}.

\bibitem[Jha et~al.(2022)Jha, Gong, Wang, Turner, and Yao]{jha2022neural}
Jha, S., Gong, D., Wang, X., Turner, R.~E., and Yao, L.
\newblock The neural process family: Survey, applications and perspectives.
\newblock \emph{arXiv preprint arXiv:2209.00517}, 2022.

\bibitem[Khan et~al.(2022)Khan, Naseer, Hayat, Zamir, Khan, and Shah]{khan2022transformers}
Khan, S., Naseer, M., Hayat, M., Zamir, S.~W., Khan, F.~S., and Shah, M.
\newblock Transformers in vision: A survey.
\newblock \emph{ACM computing surveys (CSUR)}, 54\penalty0 (10s):\penalty0 1--41, 2022.

\bibitem[Kim et~al.(2019)Kim, Mnih, Schwarz, Garnelo, Eslami, Rosenbaum, Vinyals, and Teh]{kim2019attentive}
Kim, H., Mnih, A., Schwarz, J., Garnelo, M., Eslami, A., Rosenbaum, D., Vinyals, O., and Teh, Y.~W.
\newblock Attentive neural processes.
\newblock In \emph{International Conference on Learning Representations}, 2019.

\bibitem[Kumar et~al.(2019)Kumar, Zhang, and Leskovec]{kumar2019predicting}
Kumar, S., Zhang, X., and Leskovec, J.
\newblock Predicting dynamic embedding trajectory in temporal interaction networks.
\newblock In \emph{Proceedings of the 25th ACM SIGKDD International Conference on Knowledge Discovery \& Data Mining}, pp.\  1269--1278. ACM, 2019.

\bibitem[Lee et~al.(2019)Lee, Lee, Kim, Kosiorek, Choi, and Teh]{lee2019set}
Lee, J., Lee, Y., Kim, J., Kosiorek, A., Choi, S., and Teh, Y.~W.
\newblock Set transformer: A framework for attention-based permutation-invariant neural networks.
\newblock In \emph{International Conference on Machine Learning}, pp.\  3744--3753. PMLR, 2019.

\bibitem[Lee et~al.(2020)Lee, Lee, Kim, Yang, Hwang, and Teh]{lee2020bootstrapping}
Lee, J., Lee, Y., Kim, J., Yang, E., Hwang, S.~J., and Teh, Y.~W.
\newblock Bootstrapping neural processes.
\newblock \emph{Advances in neural information processing systems}, 33:\penalty0 6606--6615, 2020.

\bibitem[Li et~al.(2022{\natexlab{a}})Li, Karaman, and Sze]{zhi2022memory}
Li, P. Z.~X., Karaman, S., and Sze, V.
\newblock Memory-efficient gaussian fitting for depth images in real time.
\newblock In \emph{2022 International Conference on Robotics and Automation (ICRA)}, pp.\  8003--8009, 2022{\natexlab{a}}.
\newblock \doi{10.1109/ICRA46639.2022.9811682}.

\bibitem[Li et~al.(2022{\natexlab{b}})Li, Gao, Ziesche, and Neumann]{li2022category}
Li, Y., Gao, N., Ziesche, H., and Neumann, G.
\newblock Category-agnostic 6d pose estimation with conditional neural processes.
\newblock \emph{arXiv preprint arXiv:2206.07162}, 2022{\natexlab{b}}.

\bibitem[Liang \& Gao(2022)Liang and Gao]{liang2022neural}
Liang, H. and Gao, J.
\newblock How neural processes improve graph link prediction.
\newblock In \emph{ICASSP 2022-2022 IEEE International Conference on Acoustics, Speech and Signal Processing (ICASSP)}, pp.\  3543--3547. IEEE, 2022.

\bibitem[Lin et~al.(2022)Lin, Wang, Liu, and Qiu]{lin2022survey}
Lin, T., Wang, Y., Liu, X., and Qiu, X.
\newblock A survey of transformers.
\newblock \emph{AI Open}, 2022.

\bibitem[Lin et~al.(2021)Lin, Wu, Zhou, Pan, Cao, and Wang]{lin2021task}
Lin, X., Wu, J., Zhou, C., Pan, S., Cao, Y., and Wang, B.
\newblock Task-adaptive neural process for user cold-start recommendation.
\newblock In \emph{Proceedings of the Web Conference 2021}, pp.\  1306--1316, 2021.

\bibitem[Liu et~al.(2015)Liu, Luo, Wang, and Tang]{liu2015faceattributes}
Liu, Z., Luo, P., Wang, X., and Tang, X.
\newblock Deep learning face attributes in the wild.
\newblock In \emph{Proceedings of International Conference on Computer Vision (ICCV)}, December 2015.

\bibitem[Mishra et~al.(2018)Mishra, Rohaninejad, Chen, and Abbeel]{mishra2018simple}
Mishra, N., Rohaninejad, M., Chen, X., and Abbeel, P.
\newblock A simple neural attentive meta-learner.
\newblock In \emph{International Conference on Learning Representations}, 2018.

\bibitem[Nguyen \& Grover(2022)Nguyen and Grover]{nguyen2022transformer}
Nguyen, T. and Grover, A.
\newblock Transformer neural processes: Uncertainty-aware meta learning via sequence modeling.
\newblock In \emph{International Conference on Machine Learning}, pp.\  16569--16594. PMLR, 2022.

\bibitem[Peng et~al.(2023)Peng, Alcaide, Anthony, Albalak, Arcadinho, Biderman, Cao, Cheng, Chung, Derczynski, Du, Grella, GV, He, Hou, Kazienko, Kocon, Kong, Koptyra, Lau, Lin, Mantri, Mom, Saito, Song, Tang, Wind, Wo{\'z}niak, Zhang, Zhou, Zhu, and Zhu]{peng2023rwkv}
Peng, B., Alcaide, E., Anthony, Q.~G., Albalak, A., Arcadinho, S., Biderman, S., Cao, H., Cheng, X., Chung, M.~N., Derczynski, L., Du, X., Grella, M., GV, K.~K., He, X., Hou, H., Kazienko, P., Kocon, J., Kong, J., Koptyra, B., Lau, H., Lin, J., Mantri, K. S.~I., Mom, F., Saito, A., Song, G., Tang, X., Wind, J.~S., Wo{\'z}niak, S., Zhang, Z., Zhou, Q., Zhu, J., and Zhu, R.-J.
\newblock {RWKV}: Reinventing {RNN}s for the transformer era.
\newblock In \emph{The 2023 Conference on Empirical Methods in Natural Language Processing}, 2023.
\newblock URL \url{https://openreview.net/forum?id=7SaXczaBpG}.

\bibitem[Rabe \& Staats(2022)Rabe and Staats]{rabe2022self}
Rabe, M.~N. and Staats, C.
\newblock Self-attention does not need $o(n^{2})$ memory.
\newblock \emph{arXiv preprint arXiv:2112.05682}, 2022.

\bibitem[Requeima et~al.(2019)Requeima, Gordon, Bronskill, Nowozin, and Turner]{requeima2019fast}
Requeima, J., Gordon, J., Bronskill, J., Nowozin, S., and Turner, R.~E.
\newblock Fast and flexible multi-task classification using conditional neural adaptive processes.
\newblock \emph{Advances in Neural Information Processing Systems}, 32, 2019.

\bibitem[Riquelme et~al.(2018)Riquelme, Tucker, and Snoek]{riquelme2018deep}
Riquelme, C., Tucker, G., and Snoek, J.
\newblock Deep bayesian bandits showdown: An empirical comparison of bayesian deep networks for thompson sampling.
\newblock \emph{arXiv preprint arXiv:1802.09127}, 2018.

\bibitem[Shchur et~al.(2020)Shchur, Biloš, and Günnemann]{Shchur2020Intensity-Free}
Shchur, O., Biloš, M., and Günnemann, S.
\newblock Intensity-free learning of temporal point processes.
\newblock In \emph{International Conference on Learning Representations}, 2020.
\newblock URL \url{https://openreview.net/forum?id=HygOjhEYDH}.

\bibitem[Singh et~al.(2019)Singh, Yoon, Son, and Ahn]{singh2019sequential}
Singh, G., Yoon, J., Son, Y., and Ahn, S.
\newblock Sequential neural processes.
\newblock \emph{Advances in Neural Information Processing Systems}, 32, 2019.

\bibitem[Tambon et~al.(2022)Tambon, Laberge, An, Nikanjam, Mindom, Pequignot, Khomh, Antoniol, Merlo, and Laviolette]{tambon2022certify}
Tambon, F., Laberge, G., An, L., Nikanjam, A., Mindom, P. S.~N., Pequignot, Y., Khomh, F., Antoniol, G., Merlo, E., and Laviolette, F.
\newblock How to certify machine learning based safety-critical systems? a systematic literature review.
\newblock \emph{Automated Software Engineering}, 29\penalty0 (2):\penalty0 38, 2022.

\bibitem[Vaswani et~al.(2017)Vaswani, Shazeer, Parmar, Uszkoreit, Jones, Gomez, Kaiser, and Polosukhin]{vaswani2017attention}
Vaswani, A., Shazeer, N., Parmar, N., Uszkoreit, J., Jones, L., Gomez, A.~N., Kaiser, {\L}., and Polosukhin, I.
\newblock Attention is all you need.
\newblock \emph{Advances in neural information processing systems}, 30, 2017.

\bibitem[Vaughan et~al.(2021)Vaughan, Tebbutt, Hosking, and Turner]{vaughan2021convolutional}
Vaughan, A., Tebbutt, W., Hosking, J.~S., and Turner, R.~E.
\newblock Convolutional conditional neural processes for local climate downscaling.
\newblock \emph{arXiv preprint arXiv:2101.07950}, 2021.

\bibitem[Willi et~al.(2019)Willi, Masci, Schmidhuber, and Osendorfer]{willi2019recurrent}
Willi, T., Masci, J., Schmidhuber, J., and Osendorfer, C.
\newblock Recurrent neural processes.
\newblock \emph{arXiv preprint arXiv:1906.05915}, 2019.

\bibitem[Wu et~al.(2022)Wu, Lan, Qian, Gu, Geramifard, and Yu]{wu2022memformer}
Wu, Q., Lan, Z., Qian, K., Gu, J., Geramifard, A., and Yu, Z.
\newblock Memformer: A memory-augmented transformer for sequence modeling.
\newblock In \emph{Findings of the Association for Computational Linguistics: AACL-IJCNLP 2022}, pp.\  308--318, 2022.

\bibitem[Zaheer et~al.(2017)Zaheer, Kottur, Ravanbakhsh, Poczos, Salakhutdinov, and Smola]{zaheer2017deep}
Zaheer, M., Kottur, S., Ravanbakhsh, S., Poczos, B., Salakhutdinov, R.~R., and Smola, A.~J.
\newblock Deep sets.
\newblock \emph{Advances in neural information processing systems}, 30, 2017.

\bibitem[Zintgraf et~al.(2021)Zintgraf, Schulze, Lu, Feng, Igl, Shiarlis, Gal, Hofmann, and Whiteson]{zintgraf2021varibad}
Zintgraf, L., Schulze, S., Lu, C., Feng, L., Igl, M., Shiarlis, K., Gal, Y., Hofmann, K., and Whiteson, S.
\newblock Varibad: Variational bayes-adaptive deep rl via meta-learning.
\newblock \emph{Journal of Machine Learning Research}, 22\penalty0 (289):\penalty0 1--39, 2021.

\bibitem[Zuo et~al.(2020)Zuo, Jiang, Li, Zhao, and Zha]{zuo2020transformer}
Zuo, S., Jiang, H., Li, Z., Zhao, T., and Zha, H.
\newblock Transformer hawkes process.
\newblock In \emph{International conference on machine learning}, pp.\  11692--11702. PMLR, 2020.

\end{thebibliography}
\bibliographystyle{icml2024}

\newpage
\appendix
\onecolumn
\section{Appendix: Additional Proof Details}

In this section, we (1) provide formal proof for Cross Attention's efficient updates property, (2) derive a practical implementation of the efficient updates property that avoids numerical issues, (3) show that CMANPs satisfy context and target invariance properties, and (4) include computational complexity analysis for CMANP-AND.

\subsection{Cross Attention's Efficient Updates Proof} \label{appendix:proof:efficient_updates}

Here, we detail how $\mathrm{CrossAttention}(L, \mathcal{D}_C \cup \mathcal{D}_U)$ can be efficiently computed given $\mathrm{CrossAttention}(L, \mathcal{D}_C)$ in only $\mathcal{O}(|\mathcal{D}_U| |L|)$ computation, i.e., an amount of memory independent of the number of original context datapoints $|\mathcal{D}_C|$. For brevity, let $\mathrm{CA} = \mathrm{CrossAttention}(L, \mathcal{D}_C)$ and $\mathrm{CA'} = \mathrm{CrossAttention}(L, \mathcal{D}_C \cup \mathcal{D}_U)$.

Recall, Cross Attention is computed as follows: 
$$\mathrm{CrossAttention}(L, \mathcal{D}_C) = \mathrm{softmax}(Q K^T)V
$$
where $Q = L W_q$ is the query matrix, $K = \mathcal{D}_C W_k$ is the key matrix, and $V = \mathcal{D}_C W_v$ is the value matrix. $W_q, W_k, W_v \in \mathbb{R}^{d \times d}$ are weight matrices (learned parameters) that project the input tokens. $\mathrm{softmax}(QK^T)$ aims to compute the weight to give the respective context tokens for computing a weighted average. 

Without loss of generality, for simplicity, we first consider the $j$-th output vector of the Cross Attention ($\mathrm{CA}_j$). The value of the $j$-th output of the Cross Attention is as follows:
$$
    \mathrm{CA}_j = \sum_{i=1}^{|\mathcal{D}_C|} \frac{\exp(s_i)}{c_j} v_i
$$
where $s_i = Q_{j,:}(K_{i,:})^T$, $v_i = V_{i, :}$, and $c_j = \sum_{i=1}^{|\mathcal{D}_C|} \exp(s_i)$ (a cached normalizing constant). 

When new data points $\mathcal{D}_U$ are added to the context in the computation of $\mathrm{CA'} = \mathrm{CrossAttention}(L, \mathcal{D}_C \cup \mathcal{D}_U)$, the key and value matrices are computed as follows: $K' \leftarrow [\mathcal{D}_C, \mathcal{D}_U] W_k$ and $V' \leftarrow [\mathcal{D}_C, \mathcal{D}_U] W_v$ where $[\mathcal{D}_C, \mathcal{D}_U]$ is the stacking of the new data points $\mathcal{D}_U$ on the original context data $\mathcal{D}_C$. We can simplify the computation of the key value matrices as follows: $K' = [\mathcal{D}_C, \mathcal{D}_U] W_k = [\mathcal{D}_C W_k, \mathcal{D}_U W_k] = [K, \mathcal{D}_U W_k]$ and $V' = [\mathcal{D}_C, \mathcal{D}_U] W_v = [V, \mathcal{D}_U W_v]$. As a result, these key value matrices are simply the original key value matrices with $|\mathcal{D}_U|$ additional rows corresponding to that of the new data points.
In contrast, the query matrix is unchanged, i.e., $Q' = L W_q = Q$.

The updated Cross Attention ($\mathrm{CA'} = \mathrm{CrossAttention}(L, \mathcal{D}_C \cup \mathcal{D}_U)$) with $\mathcal{D}_U$ new data points is computed as follows: 
$$
\mathrm{CA'}_j = \sum_{i=1}^{|\mathcal{D}_C|+|\mathcal{D}_U|} \frac{\exp(s'_i)}{c'_j} v'_i =  \sum_{i=1}^{|\mathcal{D}_C|} \frac{\exp(s'_i)}{c'_j} v'_i + \sum_{i=|\mathcal{D}_C|+1}^{|\mathcal{D}_C|+|\mathcal{D}_U|} \frac{\exp(s'_i)}{c'_j} v'_i
$$
where $s'_i = Q'_{j,:}(K'_{i,:})^T$, $v'_i = V'_{i, :}$, and $c'_j = \sum_{i=1}^{|\mathcal{D}_C|+|\mathcal{D}_U|} \exp(s'_i)$ (a new normalizing constant).  

Since $K' = [K, \mathcal{D}_U W_k]$ and $Q' = Q$, thus $\forall i \in \{1, \ldots, |\mathcal{D}_C|\} \,\, s'_i = Q'_{j,:}(K'_{i,:})^T = Q_{j,:}(K_{i,:})^T = s_i$. Building on this and the fact that $c_j = \sum_{i=1}^{|\mathcal{D}_C|} \exp(s_i)$, we also have $c'_j = \sum_{i=1}^{|\mathcal{D}_C|+|\mathcal{D}_U|} \exp(s'_i) = c_j + \sum_{i=|\mathcal{D}_C|+1}^{|\mathcal{D}_C|+|\mathcal{D}_U|} \exp(s'_i)$, i.e., $c'_j$ can be computed from $c_j$ via a rolling sum in $\mathcal{O}(|\mathcal{D}_U|)$.

Furthermore, since $V' = [V, \mathcal{D}_U W_v]$, we also have $\forall i \in \{1, \ldots, |\mathcal{D}_C|\} \,\, v'_i = V'_{i, :} = V_{i, :} = v_i$. 

Combining the prior derivations, we can re-write the updated $\mathrm{CA'}_j$ as a rolling average that only requires $\mathcal{O}(|\mathcal{D}_U|)$ compute:
\begin{align*}
    \mathrm{CA'}_j &= \sum_{i=1}^{|\mathcal{D}_C|} \frac{\exp(s_i)}{c'_j} v_i + \sum_{i=|\mathcal{D}_C|+1}^{|\mathcal{D}_C|+|\mathcal{D}_U|} \frac{\exp(s'_i)}{c'_j} v'_i \\
    &= \frac{c_j}{c'_j} \sum_{i=1}^{|\mathcal{D}_C|} \frac{\exp(s_i)}{c_j} v_i + \sum_{i=|\mathcal{D}_C|+1}^{|\mathcal{D}_C|+|\mathcal{D}_U|} \frac{\exp(s'_i)}{c'_j} v'_i  \\ 
    &= \frac{c_j}{c'_j} \mathrm{CA}_j + \sum_{i=|\mathcal{D}_C|+1}^{|\mathcal{D}_C|+|\mathcal{D}_U|} \frac{\exp(s'_i)}{c'_j} v'_i  \\ 
\end{align*}

where $c'_j = c_j + \sum_{i=|\mathcal{D}_C|+1}^{|\mathcal{D}_C|+|\mathcal{D}_U|} \exp(s'_i)$ computed via a $\mathcal{O}(|\mathcal{D}_U|)$ computation rolling sum. 

Due to the space limitations of the main paper, we aim to simplify the notation.
Note that $v_i$ and $s_i$ were only defined for $i \in \{1, \ldots, |\mathcal{D}_C|\}$. Furthermore, $\forall i \in \{1, \ldots, |\mathcal{D}_C|\} \,\, v'_i = v_i$ and $s'_i = s_i$. As such, we abuse the notation and simply replace $s'_i$ and $v'_i$ with $s_i$ and $v_i$, resulting in the following formulation seen in the main paper:

$$
\mathrm{CA'}_j = \frac{c_j}{c'_j} \mathrm{CA}_j + \sum_{i=|\mathcal{D}_C|+1}^{|\mathcal{D}_C|+|\mathcal{D}_U|} \frac{\exp(s_i)}{c'_j} v_i  \\ 
$$

There are $L$ queries, so computing the updated output $\mathrm{CA'}$ given $\mathrm{CA}$ uses $\mathcal{O}(|\mathcal{D}_U| |L|)$ computation in total.

\textbf{Practical Implementation:}

In practice, the aforementioned implementation is not stable due to requiring the computation of $c'_j$ from $c_j$. 
Specifically, $c'_j = c_j + \sum_{i=|\mathcal{D}_C|+1}^{|\mathcal{D}_C|+|\mathcal{D}_U|} \exp(s_i)$ runs into underflow and overflow problems due to being a sum of exponentials. As such, we derive a different practical formulation to avoid numerical issues.

In the practical formulation, instead of computing $\mathrm{CA'}$ from $\mathrm{CA}$ using $C'$ and $C$, we can instead use $\log(C')$ and $\log(C)$ as follows:
\begin{align*}
    \mathrm{CA'_j} &= \exp(\log(c_j) - \log(c'_j)) \times \mathrm{CA_j} + \sum_{i=|\mathcal{D}_C|+1}^{|\mathcal{D}_C|+|\mathcal{D}_U|} \exp(s_i - \log(c'_j)) v_i
\end{align*}

$\log(c'_j)$ is computed from $\log(c_j)$ in a rolling sum manner: $\log(c'_j) = \log(c_j) + \mathrm{softplus}(t_j)$ where $t_j = \log(\sum_{i=|\mathcal{D}_C|+1}^{|\mathcal{D}_C|+|\mathcal{D}_U|} \exp(s_i - \log(c_j)))$. $t_j$ can be computed accurately using the log-sum-exp trick. This method of implementation avoids the underflow and overflow issue while still computing the same updated Cross Attention result $CA'$. We detail the derivation of this practical implementation below:

\textbf{Practical Implementation (Derivation): }

\begin{align*}
    c_j &= \sum_{i=1}^{|\mathcal{D}_C|} \exp(s_i) \\
    c'_j &= \sum_{i=1}^{|\mathcal{D}_C|+|\mathcal{D}_U|} \exp(s_i) \\
    &= \sum_{i=1}^{|\mathcal{D}_C|} \exp(s_i) + \sum_{i=|\mathcal{D}_C|+1}^{|\mathcal{D}_C|+|\mathcal{D}_U|} \exp(s_i) \\
\end{align*}

\begin{align*}
    \log(c'_j) - \log(c_j) &= \log(\sum_{i=1}^{|\mathcal{D}_C|+|\mathcal{D}_U|} \exp(s_i)) - \log(\sum_{i=1}^{|\mathcal{D}_C|} \exp(s_i)) \\
    \log(c'_j) &= \log(c_j) + \log(\frac{\sum_{i=1}^{|\mathcal{D}_C|+|\mathcal{D}_U|} \exp(s_i)}{\sum_{i=1}^{|\mathcal{D}_C|} \exp(s_i)}) \\
    \log(c'_j) &= \log(c_j) + \log(1 + \frac{\sum_{i=|\mathcal{D}_C|+1}^{|\mathcal{D}_C|+|\mathcal{D}_U|} \exp(s_i)}{\sum_{i=1}^{|\mathcal{D}_C|} \exp(s_i)}) \\
    \log(c'_j) &= \log(c_j) + \log(1 + \frac{\sum_{i=|\mathcal{D}_C|+1}^{|\mathcal{D}_C|+|\mathcal{D}_U|} \exp(s_i)}{\exp(\log(c_j))}) \\
    \log(c'_j) &= \log(c_j) + \log(1 + \sum_{i=|\mathcal{D}_C|+1}^{|\mathcal{D}_C|+|\mathcal{D}_U|} \exp(s_i - \log(c_j))) \\
\end{align*}
Let $t_j = \log(\sum_{i=|\mathcal{D}_C|+1}^{|\mathcal{D}_C|+|\mathcal{D}_U|} \exp(s_i - \log(c_j)))$. Note that $t_j$ can be computed efficiently and accurately using the log-sum-exp trick in $\mathcal{O}(|\mathcal{D}_U|)$. Also, recall the $\mathrm{softplus}$ function is defined as follows: $\mathrm{softplus}(k) = \log(1+ \exp(k))$. Leveraging these, we have the following:

\begin{align*}
    \log(c'_j) &= \log(c_j) + \log(1 + \exp(t_j)) \\
    &= \log(c_j) + \mathrm{softplus}(t_j)
\end{align*}
Now recall the original update formulation:
$$
    \mathrm{CA'}_j = \frac{c_j}{c'_j} \times \mathrm{CA}_j \,\,+ \sum_{i=|\mathcal{D}_C|+1}^{|\mathcal{D}_C|+|\mathcal{D}_U|} \frac{\exp(s_i)}{c'_j} v_i
$$
Re-formulating it using $\log(C)$ and $\log(C')$ instead of $C$ and $C'$ we have the following update formulation:
$$
    \mathrm{CA'}_j = \exp(\log(c_j) - \log(c'_j)) \times \mathrm{CA}_j + \sum_{i=|\mathcal{D}_C|+1}^{|\mathcal{D}_C|+|\mathcal{D}_U|} \exp(s_i - \log(c'_j)) v_i
$$
where $\log(c'_j) = \log(c_j) + \log(1 + \sum_{i=|\mathcal{D}_C|+1}^{|\mathcal{D}_C|+|\mathcal{D}_U|} \exp(s_i - \log(c_j)))$ and $t_j = \log(\sum_{i=|\mathcal{D}_C|+1}^{|\mathcal{D}_C|+|\mathcal{D}_U|} \exp(s_i - \log(c_j)))$ such that $t_j$ is computed using the log-sum-exp trick. 
Notably, this formulation avoids the underflow and overflow issues while still only requiring $\mathcal{O}(|\mathcal{D}_U|)$ computation to compute $\mathrm{CA'}_j$.

\subsection{Additional Properties}

In this section, we show that CMANPs uphold the context and target invariance properties. 

\textbf{Property: Context Invariance.} A Neural Process $p_\theta$ is context invariant if for any choice of permutation function $\pi$, context data points $\{(x_i, y_i)\}_{i=1}^{N}$, and target data points $x_{N+1:N+M}$, 
\begin{align*}
p_\theta(y_{N+1:N+M} | x_{N+1:N+M} , x_{1:N}, y_{1:N}) = 
p_\theta(y_{N+1:N+M} | x_{N+1:N+M} , x_{\pi(1):\pi(N)}
, y_{\pi(1):\pi(N)})
\end{align*}

\textbf{Proof Outline:} Since CMANPs retrieve information from compressed encodings of the context dataset $\mathcal{D}_C$ computed using CMABs (Constant Memory Attention Blocks). It suffices to show that CMABs compute their output while being order invariant in the context dataset. 

In brief CMAB's work as follows:
$$
    \mathbf{CMAB}(L_I, \mathcal{D}_C) = 
    \mathbf{SA}(\mathbf{CA}(L_I, \mathbf{SA}(\mathbf{CA}(L_B, \mathcal{D}_C))))
$$

where $L_I$ is a set of vectors outputted by prior blocks, $L_B$ is a set of vectors whose values are learned during training, and $\mathcal{D_C}$ are the set of inputs in which we wish to be order invariant in.

The first cross-attention to be computed is $\mathbf{CA}(L_B, \mathcal{D}_C)$. A nice feature of cross-attention is that its order-invariant in the keys and values; in this case, these are embeddings of $\mathcal{D}_C$. In other words, the output of $\mathbf{CA}(L_B, \mathcal{D})$ is order invariant in the input data $\mathcal{D}$. 

The inputs to the remaining self-attention and cross-attention blocks take as input a combination of: $L_I$ and the output of $\mathbf{CA}(L_B, \mathcal{D}_C)$, both of which are order invariant in $\mathcal{D}_C$. As such, the output of CMAB is order invariant in $\mathcal{D}_C$. Therefore, CMANPs are also context invariant as required. 

\textbf{Property: Target Equivariance.} A model $p_\theta$ is target
equivariant if for any choice of permutation function $\pi$, context data points $\{(x_i, y_i)\}_{i=1}^{N}$, and target data points $x_{N+1:N+M}$, 
\begin{align*}
p_\theta(y_{N+1:N+M} | x_{N+1:N+M} , x_{1:N}, y_{1:N}) = p_\theta(y_{\pi(N+1):\pi(N+M)}
| x_{\pi(N+1):\pi(N+M)}
, x_{1:N}, y_{1:N})
\end{align*}

\textbf{Proof Outline:} The vanilla variant of CMANPs makes predictions similar to that of LBANPs~\citep{feng2023latent} by retrieving information from a set of latent vectors via cross-attention. After retrieving information, the final output is computed using an MLP (Predictor). 
The architecture design of LBANPs ensure that the result is equivalent to making the predictions independently.
Since CMANPs use the same querying mechanism as LBANPs, therefore CMANPs preserve target equivariance the same way LBANPs do. 

However, for the Autoregressive Not-Diagonal variant (CMANP-AND), the target equivariance is not held as it depends on the order in which the data points are processed. This property is in common with that of several other NP methods by \citet{nguyen2022transformer} and \citet{bruinsma2023autoregressive}.

\subsection{Complexity Analysis for CMANP-AND} \label{appendix:proof:complexity_analysis_cmanp_and}

For a batch of $M$ data points and a prediction block size of $b_Q$ (hyperparameter constant), there are $\ceil{\frac{M}{b_Q}}$ batches of data points whose predictions are made autoregressively. 
Each batch incurs a constant complexity of $\mathcal{O}(b_Q)^2$ due to predicting a full covariance matrix. 
As such for a batch of $M$ target data points, CMANP-AND requires a sub-quadratic total computation of $\mathcal{O}(\ceil{\frac{M}{b_Q}} b_Q^2) = \mathcal{O}(Mb_Q)$ with a sequential computation length of $\mathcal{O}(\frac{M}{b_Q})$.
Crucially, CMANP-AND only requires constant memory in $|\mathcal{D}_C|$ and linear memory in $M$, making it significantly more efficient than prior works which required at least quadratic memory.

\subsection{Cross Attention Efficient Updates Property is inapplicable to popular attention mechanisms}\label{appendix:proof:eff_upd_perceiver}

Cross Attention's efficient updates property allows $\mathrm{CrossAttention}(L, \mathcal{D}_C \cup \mathcal{D}_U)$ to be efficiently computed given  $\mathrm{CrossAttention}(L, \mathcal{D}_C)$. This property is conditional on the following: (1) the query tokens remaining the same and (2) the new tokens being stacked on the old context tokens. However, this is not the case for popular attention methods. 

For example, the closest to our work is the family of methods based on Perceiver~\citep{jaegle2021perceiverio, jaegle2021perceiver}. Perceiver encodes the context dataset via a series of iterative attention blocks. Each iterative attention block consists of a Cross Attention layer to retrieve information from the context dataset for a given set of latents and self-attention layer(s) applied to the latents to compute higher-order information.  As a result of stacking these layers, Perceiver fails condition (1) due to the query vectors changing while being stacked.

In brief, Perceiver's iterative attention layers work as follows:

$$
    L_i = \mathrm{SelfAttention}(\mathrm{CrossAttention}(L_{i-1}, \mathcal{D}_C))
$$
where $L_i$ is a fixed-sized set of latents computed iteratively.

When the context dataset is updated with new data $\mathcal{D}_U$, the computation is as follows:

$$
    L'_i = \mathrm{SelfAttention}(\mathrm{CrossAttention}(L'_{i-1}, \mathcal{D}_C \cup \mathcal{D}_U))
$$

To achieve the efficient updates, it requires that $\mathrm{CrossAttention}(L'_{i-1}, \mathcal{D}_C \cup \mathcal{D}_U)$ be efficiently computed from $\mathrm{CrossAttention}(L_{i-1}, \mathcal{D}_C)$. However, $L'_{i-1} \neq L_{i-1}$ for $i > 1$. 
For example, $L'_1 = \mathrm{SelfAttention}(\mathrm{CrossAttention}(L'_0, \mathcal{D}_C \cup \mathcal{D}_U))$ and $L_1 = \mathrm{SelfAttention}(\mathrm{CrossAttention}(L_0, \mathcal{D}_C \cup \mathcal{D}_U))$.

As such, the efficient updates property of Cross Attention cannot be applied to Perceiver-style architectures.

Another example is that of Transformers. In Transformer Encoders self-attention is repeatedly applied starting from the context tokens, resulting in the outputs of the layers changing drastically when new tokens are added to the context, failing property (2). In the case of the decoder, the argument is the same as Perceiver where the stacked layers cause the queries to change overtime.

\section{Appendix: Intuition for Constant Memory Attention Block's construction} \label{appendix:cmab_intuition}

In this work, we aim to leverage an efficient attention block for Neural Processes that achieves competitive performance with prior state-of-the-art. As such, the attention block has the following requirements:

\begin{enumerate}
    \item \textbf{Computational Efficiency.} Prior attention-based models require linear or quadratic memory in the number of tokens, limiting their effectiveness in low-resource scenarios. Furthermore, they often require re-computing from scratch when receiving new tokens, i.e., expensive updates. Improving on this, we desire a constant memory version of attention that also allows for constant computation updates.
    \item \textbf{Permutation invariant in the context.} Neural Processes (NPs) (Garnelo et al., 2018) are permutation invariant in the context.
    \item \textbf{Stackable.} Modern deep learning leverages the stacking of the same kind of modules (e.g., Transformers) to construct deep models and achieve strong performance.
    \item \textbf{Computes higher-order information between tokens.} A powerful feature of Transformers that allows it to achieve strong performance.
\end{enumerate}

In Section \ref{sec:method:cross_att}, we showed that Cross Attention with a fixed query vector can compute its output in constant memory and can compute updates in constant computation. Notably, this solves the first requirement (Computational Efficiency). 

However, since previous attention-based models do not have a fixed query, this efficiency property does not apply to them. As such, we design our own attention block to address these requirements. We begin with a Cross Attention module learned with a fixed query, i.e., $L_B' \leftarrow \mathrm{CrossAttention}(L_B, \mathcal{D}_C)$ where $L_B$ is the fixed query and $\mathcal{D}_C$ is the context data. Since attention is by default permutation-invariant in the context, this module resolves the second requirement (Permutation Invariant), i.e., $L_B'$ is permutation invariant in $\mathcal{D}_C$. 

However, this module by itself is not stackable since it comprises of only two inputs: $L_B$ the fixed query (i.e., a learned constant) and $\mathcal{D}_C$ is the context data. 
Thus to achieve the third requirement (Stackable), we introduce another cross attention block with a new input latent $L_I$ as follows: $L_I^{i+1} \leftarrow \mathrm{CrossAttention}(L_I^{i}, L_B')$ where $L_I^{i}$ is the output of the previous block and $L_I^{i+1}$ is the output of the current block. 

Finally, to achieve the fourth requirement (Higher-order information), we wrap the CrossAttention modules with a SelfAttention module.  Together, these modules result in the final version of the Constant Memory Attention Block which we proposed in the paper:

$$
L_B' 
\leftarrow 
\mathrm{SelfAttention(CrossAttention}(L_B, \mathcal{D}_C))
$$

$$
L_I^{i+1} 
\leftarrow 
\mathrm{SelfAttention(CrossAttention}(L_I^{i}, L_B'))
$$

\section{Appendix: Additional Experiments and Analyses}

\subsection{Applying CMABs to Temporal Point Processes (TPPs)}

Constant Memory Attention Blocks (1) do not leverage any modality-specific components and (2) are permutation invariant by default like Transformers. As such, CMABs appear to be naturally applicable to a broad range of applications beyond that of Neural Processes. 
As a proof of concept, we showcase the efficacy of CMABs on next-event prediction (Temporal Point Processes (TPPs)).
In brief, Temporal Point Processes are stochastic processes composed of a time series of discrete events. 
Recent works have proposed to model this via a neural network. 
Notably, models such as THP~\citep{zuo2020transformer} encode the history of past events to predict the next event, i.e., modelling the predictive distribution of the next event $p_\theta(\tau_{l+1} | \tau_{\leq l})$ where $\theta$ are the parameters of the model, $\tau$ represents an event, and $l$ is the number of events that have passed.
Typically, an event comprises a discrete temporal (time) stamp and a mark (categorical class). 

\subsubsection{Constant Memory Hawkes Processes (CMHPs)}

\begin{figure*}[h]
    \centering
    \includegraphics[width=\textwidth]{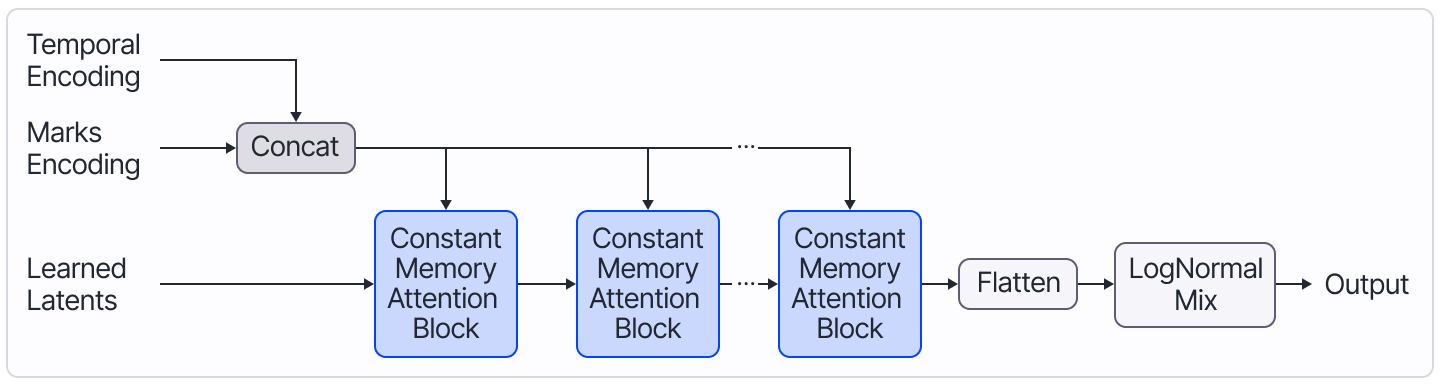}
    \caption{Constant Memory Hawkes Processes}
    \label{appendix:fig:cmhps}
\end{figure*}

Building on CMABs, we introduce the Constant Memory Hawkes Process (CMHPs) (Figure \ref{appendix:fig:cmhps}), a model which replaces the transformer layers in Transformer Hawkes Process (THP)~\citep{zuo2020transformer} with Constant Memory Attention Blocks. 
Unlike THPs which summarise the information for prediction in a single vector, CMHPs summarise it as a set of latent vectors. 
As such, a flattening operation is added at the end of the model.
Following prior work~\citep{bae2023meta,Shchur2020Intensity-Free}, the predictive distribution for THPs and CMHPs is a mixture of log-normal distribution.

\subsubsection{CMHPs: Experiments}

In these experiments, we compare CMHPs against THPs on standard TPP datasets: Mooc and Reddit. 

\textbf{Mooc Dataset}~\citep{kumar2019predicting} comprises of $7,047$ sequences. Each sequence contains the action times of an individual user of an online Mooc course with 98 categories for the marks.

\textbf{Reddit Dataset}~\citep{kumar2019predicting} comprises of $10,000$ sequences. Each sequence contains the action times from the most active users with marks being one of the $984$ the subreddit categories of each sequence.

The results (Table \ref{appendix:table:tpp}) suggest that replacing the transformer layer with CMAB (Constant Memory Attention Block) achieves comparable performance. Crucially, CMHP only requires constant memory unlike the quadratic memory required by that of THP.
Furthermore, CMHP efficiently updates its model with new data as it arrives over time which is typical in event sequence data, making it significantly more efficient than THP.

\begin{table}[]
\centering
\begin{tabular}{|l|ccc|ccc|}
\hline
\multicolumn{1}{|c|}{} & \multicolumn{3}{c|}{Mooc}                                                                                              & \multicolumn{3}{c|}{Reddit}                                                                                           \\ \hline
\multicolumn{1}{|c|}{} & \multicolumn{1}{c|}{RMSE}                    & \multicolumn{1}{c|}{NLL}                      & ACC                     & \multicolumn{1}{c|}{RMSE}                    & \multicolumn{1}{c|}{NLL}                     & ACC                     \\ \hline
THP                   & \multicolumn{1}{c|}{0.202 ± 0.017}          & \multicolumn{1}{c|}{\textbf{0.267 ± 0.164}}  & \textbf{0.336 ± 0.007} & \multicolumn{1}{c|}{\textbf{0.238 ± 0.028}} & \multicolumn{1}{c|}{\textbf{0.268 ± 0.098}} & \textbf{0.610 ± 0.002} \\ \hline
CMHP                 & \multicolumn{1}{c|}{\textbf{0.168 ± 0.011}} & \multicolumn{1}{c|}{\textbf{-0.040 ± 0.620}} & 0.237 ± 0.024          & \multicolumn{1}{c|}{\textbf{0.262 ± 0.037}} & \multicolumn{1}{c|}{\textbf{0.528 ± 0.209}} & \textbf{0.609 ± 0.003} \\ \hline
\end{tabular}
\caption{Temporal Point Processes Experiments.}
\label{appendix:table:tpp}
\end{table}

\subsection{CMANPs Experiment: Contextual Bandits}

In the Contextual Bandit setting introduced by \cite{riquelme2018deep}, a unit circle is divided into 5 sections which contain 1 low reward section and 4 high reward sections
$\delta$ defines the size of the low reward section while the 4 high reward sections have equal sizes.
In each round, the agent has to select $1$ of $5$ arms that each represent one of the regions. 
For context during the selection, the agent is given a 2-D coordinate $X$ and the actions it selected and rewards it received in previous rounds. 

If $|| X || < \delta$, then the agent is within the low reward section.
If the agent pulls arm $1$, then the agent receives a reward of $r \sim \mathcal{N}(1.2, 0.012)$. Otherwise, if the agent pulls a different arm, then it receives a reward $r \sim \mathcal{N}(1.0, 0.012)$. 
Consequently, if $||X|| \geq \delta$, then the agent is within one of the four high-reward sections.
If the agent is within a high reward region and selects the corresponding arm to the region, then the agent receives a large reward of $N \sim \mathcal{N}(50.0, 0.012)$. Alternatively, pulling arm $1$ will reward the agent with a small reward of $r \sim \mathcal{N}(1.2, 0.012)$. Pulling any of the other $3$ arms rewards the agent with an even smaller reward of $r \sim \mathcal{N}(1.0, 0.012)$.

During each training iteration, $B=8$ problems are sampled. Each problem is defined by $\{\delta_i\}_{i=1}^{B}$ which are sampled according to a uniform distribution $\delta \sim \mathcal{U}(0, 1)$. 
$N=512$ points are sampled as context data points and $M=50$ points are sampled for evaluation. 
Each data point comprises of a tuple $(X, r)$ where $X$ is the coordinate and $r$ is the reward values for the $5$ arms. 
The objective of the model during training is to predict the reward values for the $5$ arms given the coordinates (context data points).

During the evaluation, the model is run for $2000$ steps.
At each step, the agent selects the arm which maximizes its UCB (Upper-Confidence Bound). After which, the agent receives the reward value corresponding to the arm.
The performance of the agent is measured by cumulative regret. 
For comparison, we evaluate the modes with varying $\delta$ values and report the mean and standard deviation for $50$ seeds.

\textbf{Results.} In Table \ref{appendix:table:cmab}, we compare CMANPs with other NP baselines. We see that CMANP achieves competitive performance with state-of-the-art for $\delta \in \{0.7, 0.9, 0.95\}$. However, the performance degrades as $\delta$ reaches extreme values close to the limit such as $0.99$ and $0.995$ -- settings that are at the extremities of the training distribution.  

\begin{table*}[]
\centering
\begin{tabular}{|c|ccccc|}
\hline
Method  & $\delta = 0.7$       & $\delta = 0.9$       & $\delta = 0.95$      & $\delta = 0.99$      & $\delta = 0.995$     \\ \hline
Uniform & 100.00 ± 1.18        & 100.00 ± 3.03        & 100.00 ± 4.16        & 100.00 ± 7.52        & 100.00 ± 8.11        \\
CNP     & 4.08 ± 0.29          & 8.14 ± 0.33          & 8.01 ± 0.40          & 26.78 ± 0.85         & 38.25 ± 1.01         \\
CANP    & 8.08 ± 9.93          & 11.69 ± 11.96        & 24.49 ± 13.25        & 47.33 ± 20.49        & 49.59 ± 17.87        \\
NP      & 1.56 ± 0.13          & 2.96 ± 0.28          & 4.24 ± 0.22          & 18.00 ± 0.42         & 25.53 ± 0.18         \\
ANP     & 1.62 ± 0.16          & 4.05 ± 0.31          & 5.39 ± 0.50          & 19.57 ± 0.67         & 27.65 ± 0.95         \\
BNP     & 62.51 ± 1.07         & 57.49 ± 2.13         & 58.22 ± 2.27         & 58.91 ± 3.77         & 62.50 ± 4.85         \\
BANP    & 4.23 ± 16.58         & 12.42 ± 29.58        & 31.10 ± 36.10        & 52.59 ± 18.11        & 49.55 ± 14.52        \\
TNP-D   & \textbf{1.18 ± 0.94}          & \textbf{1.70 ± 0.41}          & 2.55 ± 0.43          & \textbf{3.57 ± 1.22} & \textbf{4.68 ± 1.09} \\ 
LBANP    & \textbf{1.11 ± 0.36}          & \textbf{1.75 ± 0.22}          & \textbf{1.65 ± 0.23} &  6.13 ± 0.44    & 8.76 ± 0.15    \\ \hline
CMANP (Ours)   & \textbf{0.93 ± 0.12}    & \textbf{1.56 ± 0.10}    & \textbf{1.87 ± 0.32}    & 9.04 ± 0.42          & 13.02 ± 0.03         \\ \hline
\end{tabular}
\caption{Contextual Multi-Armed Bandit Experiments with varying $\delta$. Models are evaluated according to cumulative regret (lower is better). Each model is run 50 times for each value of $\delta$. }
\label{appendix:table:cmab}
\end{table*}

\subsection{Additional Analyses}

\textbf{Memory Complexity:} In Table \ref{table:memory_complexity:all_baselines}, we include a comparison of CMANPs with all NP baselines, showing that CMANPs are amongst the best in terms of memory efficiency when compared to prior NP methods. 
Notably, the methods with a similar memory complexity to CMANPs perform significantly worse in terms of performance across the various experiments (Tables \ref{table:image_completion} and \ref{table:1d_regression}). 
As such, CMANPs provide the best trade-off regarding memory and performance.

\begin{table}[]
\centering
\scalebox{1.0}{
\begin{tabular}{cccccc}
\hline
\multicolumn{1}{|c|}{}                          & \multicolumn{1}{c|}{Conditioning}     & \multicolumn{2}{c|}{Querying}                                                 & \multicolumn{2}{c|}{Updating}                                                 \\ \hline
\multicolumn{1}{|c|}{In Terms of}               & \multicolumn{1}{c|}{$|\mathcal{D}_C|$}              & \multicolumn{1}{c|}{$|\mathcal{D}_C|$}              & \multicolumn{1}{c|}{$M$}              & \multicolumn{1}{c|}{$|\mathcal{D}_C|$}              & \multicolumn{1}{c|}{$|\mathcal{D}_U|$}            \\ \hline
\multicolumn{1}{|c|}{CNP}                       & \multicolumn{1}{c|}{\checkmark}                & \multicolumn{1}{c|}{\checkmark}                & \multicolumn{1}{c|}{\halfcheckmark} & \multicolumn{1}{c|}{\checkmark}                & \multicolumn{1}{c|}{\checkmark}                \\ 
\multicolumn{1}{|c|}{CANP}                      & \multicolumn{1}{c|}{\nocheckmark}                & \multicolumn{1}{c|}{\nocheckmark}                & \multicolumn{1}{c|}{\halfcheckmark} & \multicolumn{1}{c|}{\nocheckmark}                & \multicolumn{1}{c|}{\nocheckmark}                \\ 
\multicolumn{1}{|c|}{NP}                        & \multicolumn{1}{c|}{\checkmark}                & \multicolumn{1}{c|}{\checkmark}                & \multicolumn{1}{c|}{\halfcheckmark} & \multicolumn{1}{c|}{\checkmark}                & \multicolumn{1}{c|}{\checkmark}                \\ 
\multicolumn{1}{|c|}{ANP}                       & \multicolumn{1}{c|}{\nocheckmark}                & \multicolumn{1}{c|}{\nocheckmark}                & \multicolumn{1}{c|}{\halfcheckmark} & \multicolumn{1}{c|}{\nocheckmark}                & \multicolumn{1}{c|}{\nocheckmark}                \\ 
\multicolumn{1}{|c|}{BNP}                       & \multicolumn{1}{c|}{\checkmark}                & \multicolumn{1}{c|}{\checkmark}                & \multicolumn{1}{c|}{\halfcheckmark} & \multicolumn{1}{c|}{\checkmark}                & \multicolumn{1}{c|}{\checkmark}                \\ 
\multicolumn{1}{|c|}{BANP}                      & \multicolumn{1}{c|}{\nocheckmark}                & \multicolumn{1}{c|}{\nocheckmark}                & \multicolumn{1}{c|}{\halfcheckmark} & \multicolumn{1}{c|}{\nocheckmark}                & \multicolumn{1}{c|}{\nocheckmark}                \\ 
\multicolumn{1}{|c|}{TNP-D}                     & \multicolumn{1}{c|}{N/A}              & \multicolumn{1}{c|}{\nocheckmark}                & \multicolumn{1}{c|}{\nocheckmark}                & \multicolumn{1}{c|}{N/A}              & \multicolumn{1}{c|}{N/A}              \\ 
\multicolumn{1}{|c|}{LBANP}                     & \multicolumn{1}{c|}{\halfcheckmark} & \multicolumn{1}{c|}{\checkmark}                & \multicolumn{1}{c|}{\halfcheckmark} & \multicolumn{1}{c|}{\halfcheckmark} & \multicolumn{1}{c|}{\halfcheckmark} \\ \hline
\multicolumn{1}{|c|}{\textbf{CMANP (Ours)}}     & \multicolumn{1}{c|}{\checkmark}                & \multicolumn{1}{c|}{\checkmark}                & \multicolumn{1}{c|}{\halfcheckmark} & \multicolumn{1}{c|}{\checkmark}                & \multicolumn{1}{c|}{\checkmark}                \\ \hline
\multicolumn{1}{l}{}                            & \multicolumn{1}{l}{}                  & \multicolumn{1}{l}{}                  & \multicolumn{1}{l}{}                  & \multicolumn{1}{l}{}                  & \multicolumn{1}{l}{}                  \\ \hline
\multicolumn{1}{|c|}{TNP-ND}                    & \multicolumn{1}{c|}{N/A}              & \multicolumn{1}{c|}{\nocheckmark}                & \multicolumn{1}{c|}{\nocheckmark}                & \multicolumn{1}{c|}{N/A}              & \multicolumn{1}{c|}{N/A}              \\ 
\multicolumn{1}{|c|}{LBANP-ND}                  & \multicolumn{1}{c|}{\halfcheckmark} & \multicolumn{1}{c|}{\halfcheckmark} & \multicolumn{1}{c|}{\nocheckmark}                & \multicolumn{1}{c|}{\halfcheckmark} & \multicolumn{1}{c|}{\halfcheckmark} \\ \hline
\multicolumn{1}{|c|}{\textbf{CMANP-AND (Ours)}} & \multicolumn{1}{c|}{\checkmark}                & \multicolumn{1}{c|}{\checkmark}                & \multicolumn{1}{c|}{\halfcheckmark} & \multicolumn{1}{c|}{\checkmark}                & \multicolumn{1}{c|}{\checkmark}                \\ \hline
\end{tabular}
}
\caption{    Comparison of Memory Complexities of Neural Processes with respect to the number of context data points $|\mathcal{D}_C|$, number of target data points in a batch $M$, and the number of new data points in an update $|\mathcal{D}_U|$.
    (Green) Checkmarks represent requiring constant memory, (Orange) half checkmarks represent requiring linear memory, and (Red) Xs represent requiring quadratic or more memory. }
    \label{table:memory_complexity:all_baselines}
\end{table}

\begin{figure*}[h]
     \centering
     \includegraphics[width=0.31\textwidth]{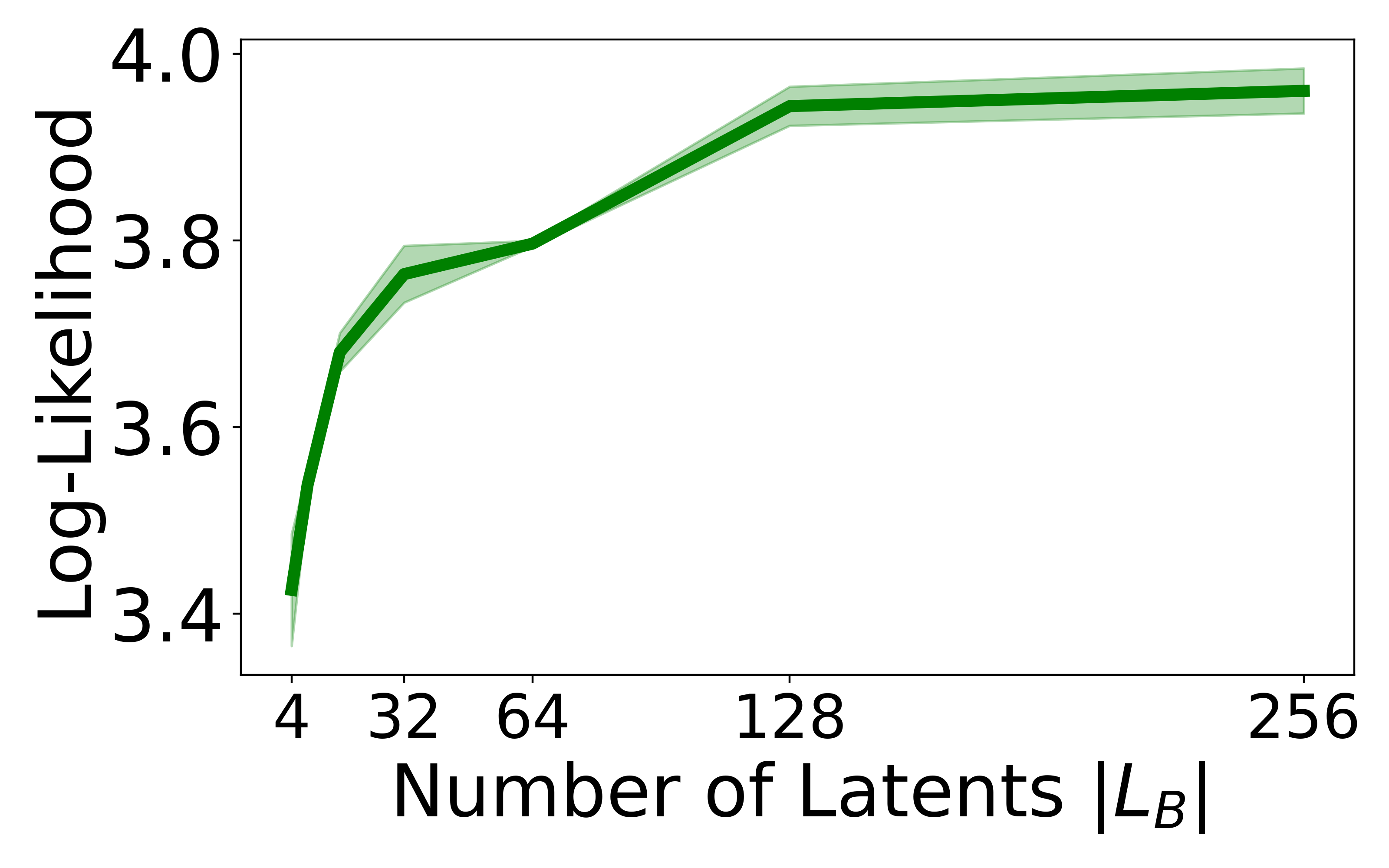}
     \includegraphics[width=0.31\textwidth]{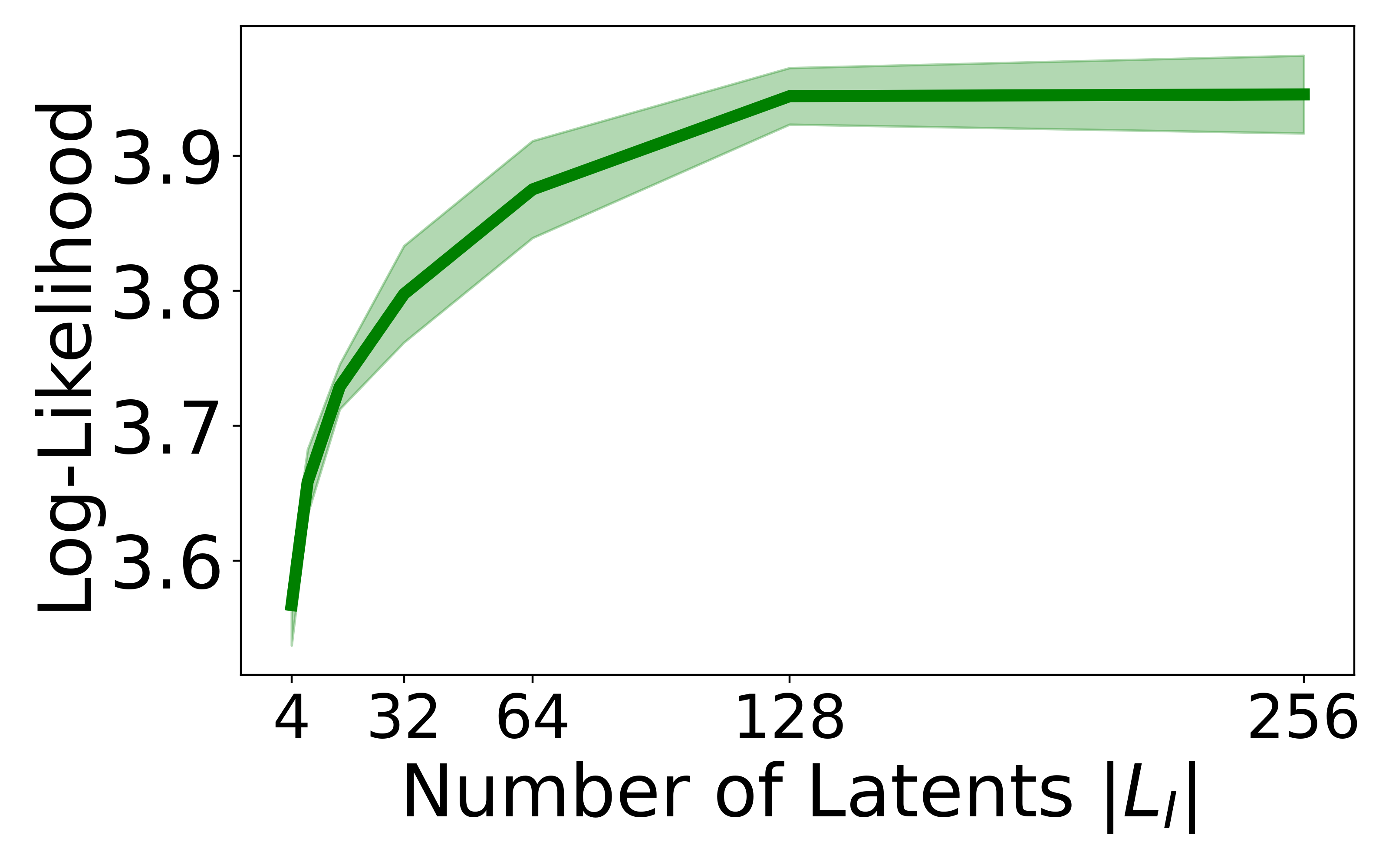}
     \includegraphics[width=0.31\textwidth]{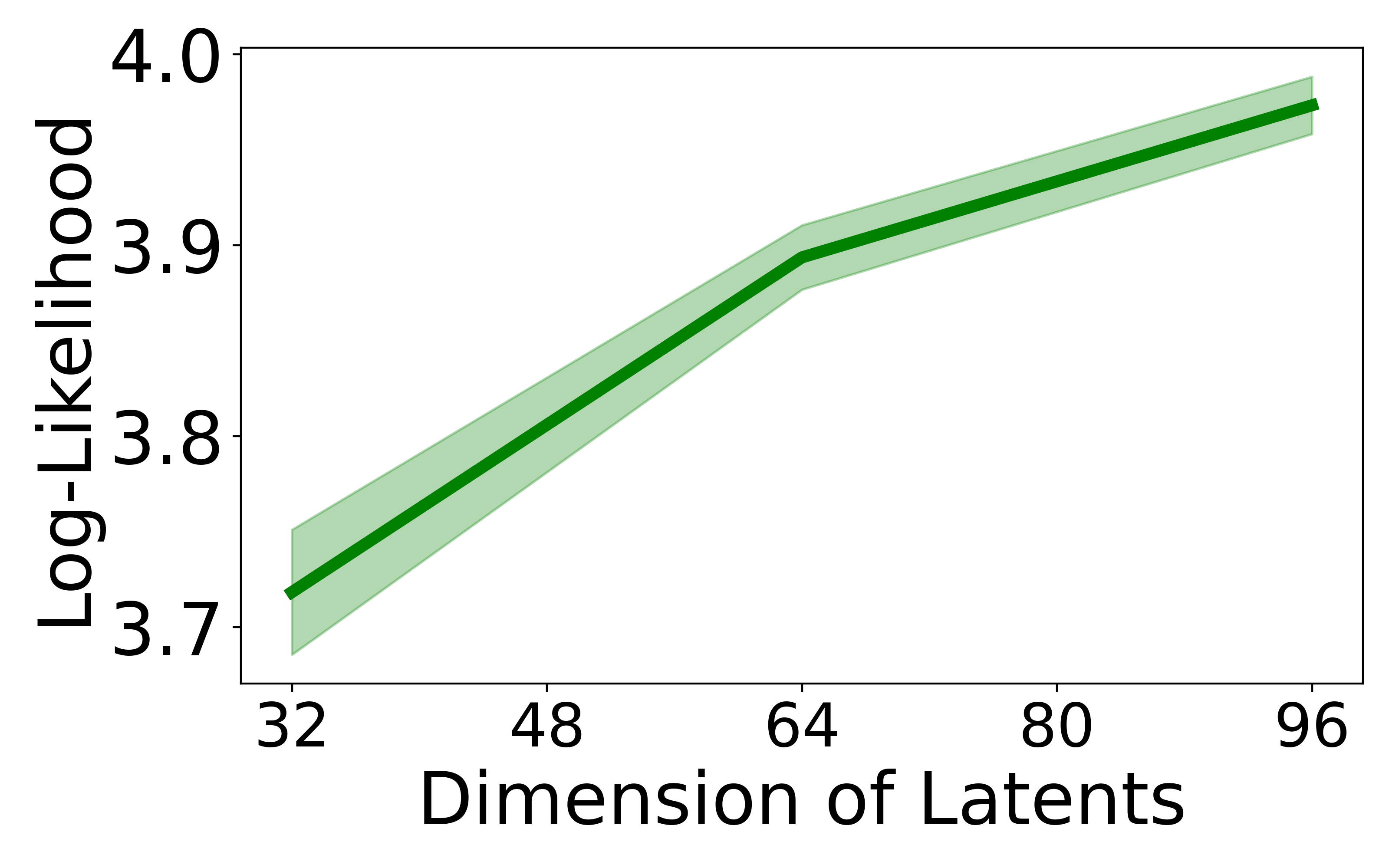}
    \caption{ CMANP's bottleneck size on performance. (Left)  $|L_B|$'s effect on performance. (Middle) $|L_I|$'s effect on performance.
    (Right) Latent dimensions' effect on performance.
    }
    \label{fig:analyses:vary_bottleneck_size}
\end{figure*}

\textbf{Effect of CMANP's bottleneck on performance:} 
As with any method that uses a bottleneck (e.g., perceiver, set transformer, LBANPs, etc...), CMAB's bottleneck usage results in some amount of information loss. Whether or not this affects the performance is dependent on (1) the amount of information loss (e.g., number of context tokens related to the bottleneck size) and (2) the intrinsic dimensionality of the task (i.e., task complexity). 

For example, in a task with a low intrinsic dimensionality, only a small amount of information from the context tokens is needed to solve it. As such, a smaller bottleneck (i.e., low values for $L_B$ and $L_I$) suffices. However, in a task with a high intrinsic dimensionality, more information is needed from the context tokens. As such, a larger bottleneck (i.e., higher values for $L_B$ and $L_I$) would be needed to hold the information.

In the main paper (Figure \ref{fig:analyses:hyperparameter_impact} (Right)), we included an analysis of the bottleneck size's effect on task performance where $L_I = L_B$. To analyze this further, we performed an additional three analyses, measuring the performance of the model with respect to individual varying values of $L_I$, $L_B$, and latent dimension. In the plots (Figure \ref{fig:analyses:vary_bottleneck_size}), we see similar results where the performance increases as the bottleneck size increases, saturating slowly. Since the performance generally increases, the bottleneck size should be selected according to the available computational resources. Regardless of the bottleneck sizes that we tried, the model achieves strong performance outperforming all non-attention based models (NP, CNP, and BNP) and several attention-based models (ANP, CANP, and BANP), making it highly applicable to low-resource scenarios.

\begin{figure}
     \centering
     \includegraphics[height=1.5in]{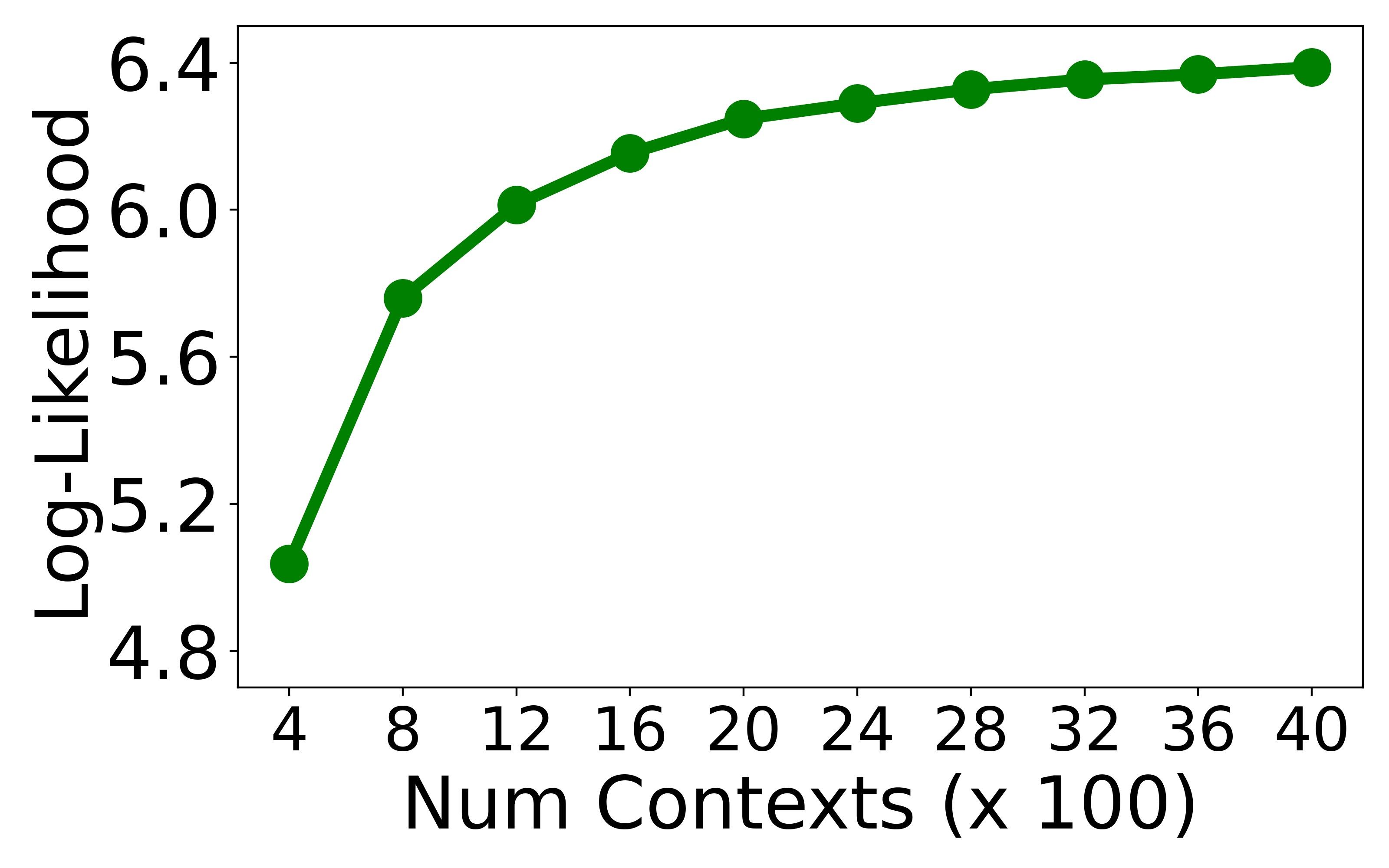}
     \caption{Number of context datapoints on performance.}
     \label{fig:analyses:vary_nContexts}
\end{figure}

To analyze the effect of a large number of contexts relative to the number of latents, we additionally run an experiment in the higher resolution setting CelebA (128x128), varying the number of contexts while fixing the number of latents to $128$. In the plot (Figure \ref{fig:analyses:vary_nContexts}), we see that the performance increases as the number of contexts increases, saturating eventually. As more pixels are being added in this experiment, the model naturally receives more relevant information for image completion, resulting in the performance increase. However, since there is ultimately a bottleneck in the model, the performance ultimately saturates given a very large number of contexts.

\begin{figure}
     \centering
     \begin{subfigure}[b]{0.45\textwidth}
         \centering
         \includegraphics[height=2.2in]{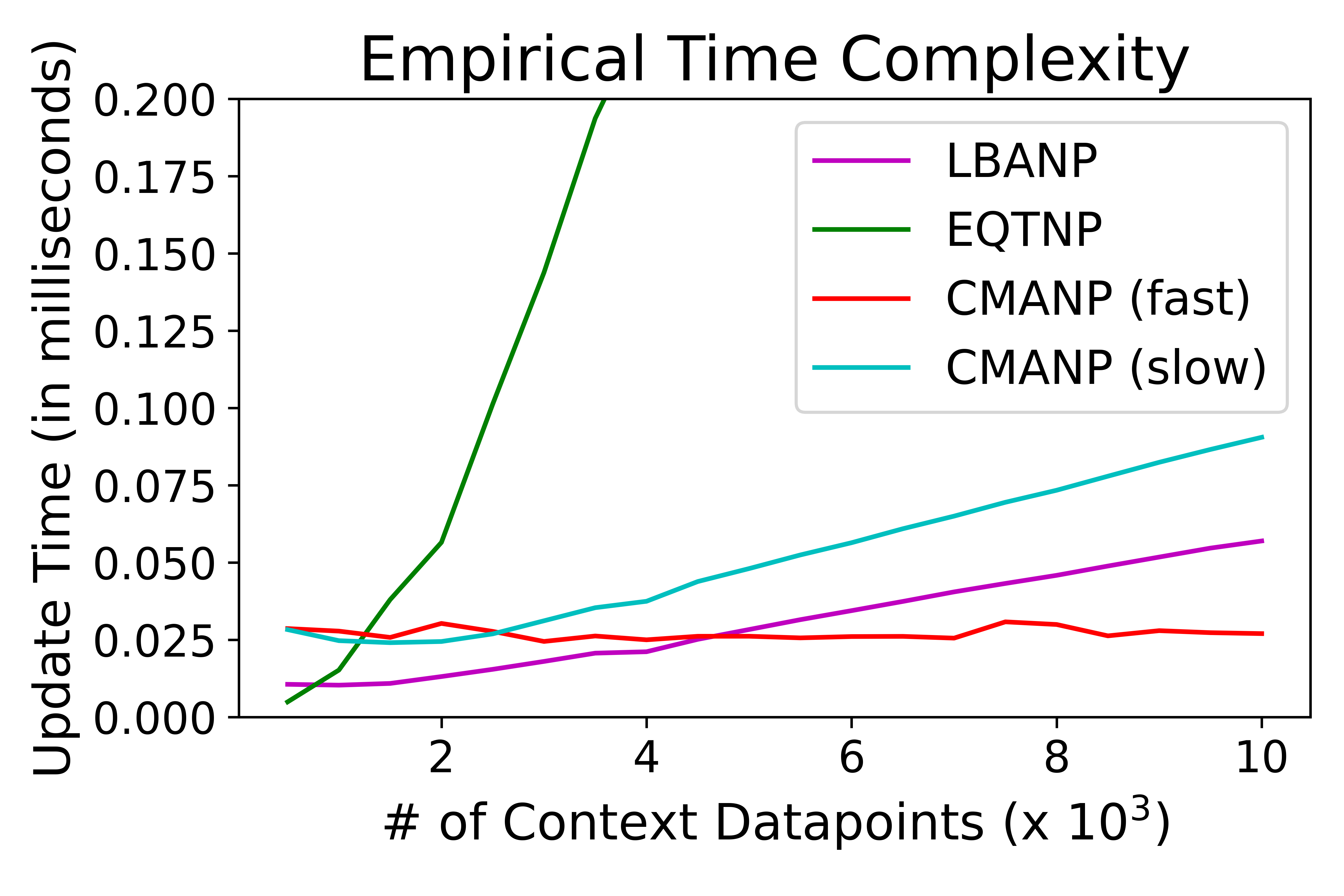}
         \caption{Empirical runtime of the update phase.}
         \label{fig:analyses:time-complexity:updates}
     \end{subfigure}
     \hfill
     \begin{subfigure}[b]{0.45\textwidth}
         \centering
         \includegraphics[height=2.2in]{imgs/empirical_time_complexity_INFERENCE_varying-contexts.png}
         \caption{Empirical runtime of the query phase.}
         \label{fig:analyses:time-complexity:inference}
     \end{subfigure}
    \caption{Analyses Graphs comparing the runtime of CMANPs with various baselines. (a) Comparison of the update procedure of CMAB-based NP (CMANPs) with Perceiver's iterative attention-based NP model (LBANPs) and a transformer-based NP model (EQTNP). CMANP (fast) refers to the CMAB's efficient update mechanism. CMANP (slow) refers to the traditional update mechanism. (b) Comparison of the query/inference process of CMANPs with LBANPs (Perceiver's iterative attention-based model), TNPs (Transformer-based model), and EQTNPs (Transformer-based model with an efficient query mechanism). }
    \label{appendix:fig:analyses:runtime}
\end{figure}

\textbf{Empirical Time Comparison with Baselines: } In Figure \ref{appendix:fig:analyses:runtime}, we compare the runtime of CMANPs with various state-of-the-art NP baselines.

In Figure \ref{fig:analyses:time-complexity:updates}, we compare the runtime of NPs' update phase. Specifically, we find that CMANPs' efficient (fast) update process based on the efficient updates property of Cross Attention only requires constant computation regardless of the number of context data points. In contrast, the traditional (slow) update process scales linearly with respect to the number of context data points. Furthermore, the Transformer-based (EQTNP) model requires quadratic time complexity and the Perceiver-based (LBANP) model requires linear time complexity. As such, CMANPs scale significantly better than prior state-of-the-art methods.

In Figure \ref{fig:analyses:time-complexity:inference}, we compare the querying (inference) runtime of CMANP with LBANPs (Perceiver's iterative attention-based model), TNPs (Transformer-based model). 
Notably, TNPs scale quadratically in runtime, making it prohibitively expensive for a large number of context data points. 
EQTNP (Efficient Queries Transformer Neural Processes) scales linearly. 
In contrast, CMANPs and LBANPs are the most efficient when performing queries since they are constant complexity (see Table \ref{table:memory_complexity:all_baselines}) regardless of the number of context data points.

\begin{figure}
     \centering
     
     \begin{subfigure}[b]{0.32\textwidth}
         \centering
         \includegraphics[width=\textwidth]{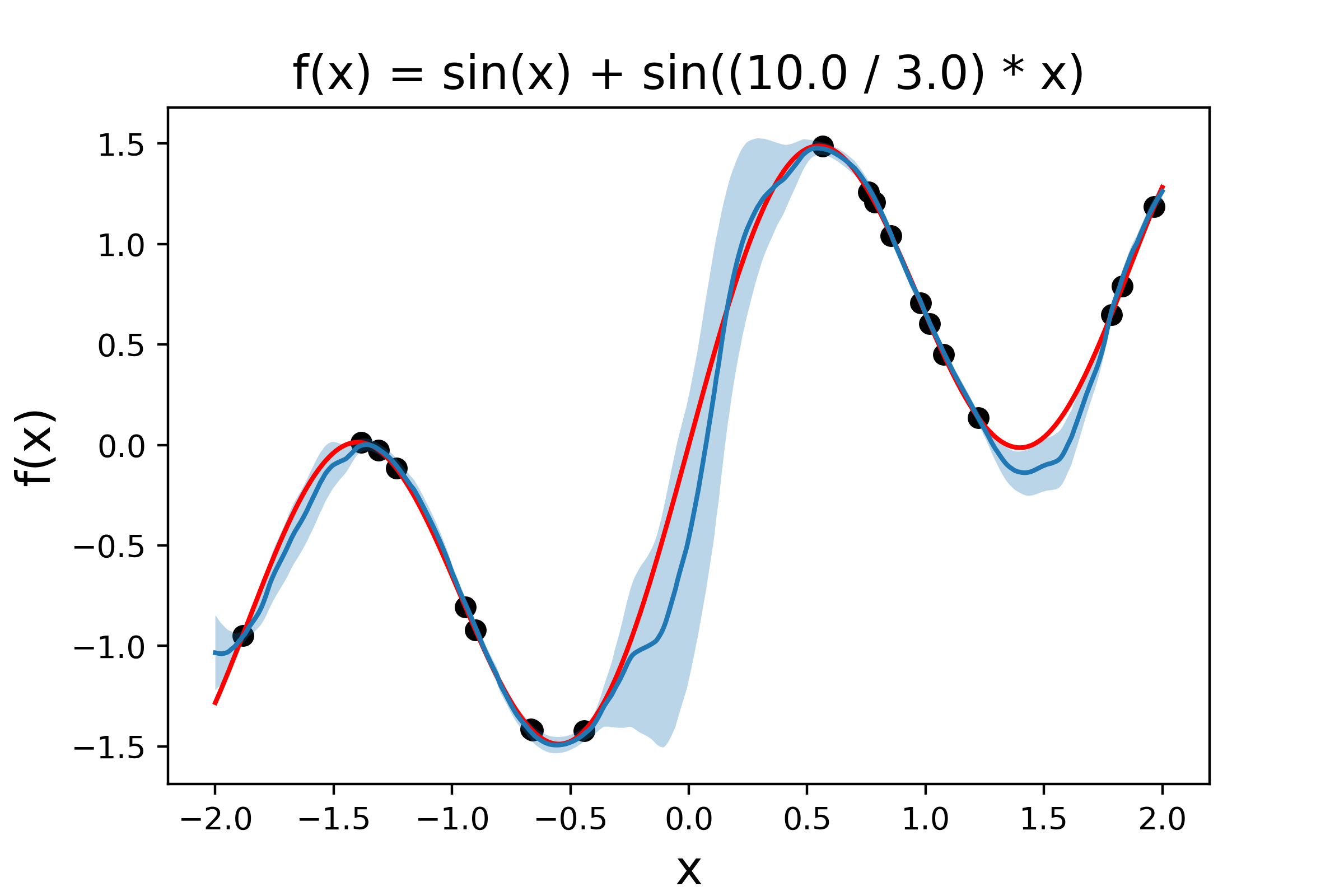}
         \caption{$y=sin(x) + sin(10/3) * x$}
        \label{fig:1-d_regression:visualization:1}
     \end{subfigure}
     \hfill
     \begin{subfigure}[b]{0.32\textwidth}
         \centering
         \includegraphics[width=\textwidth]{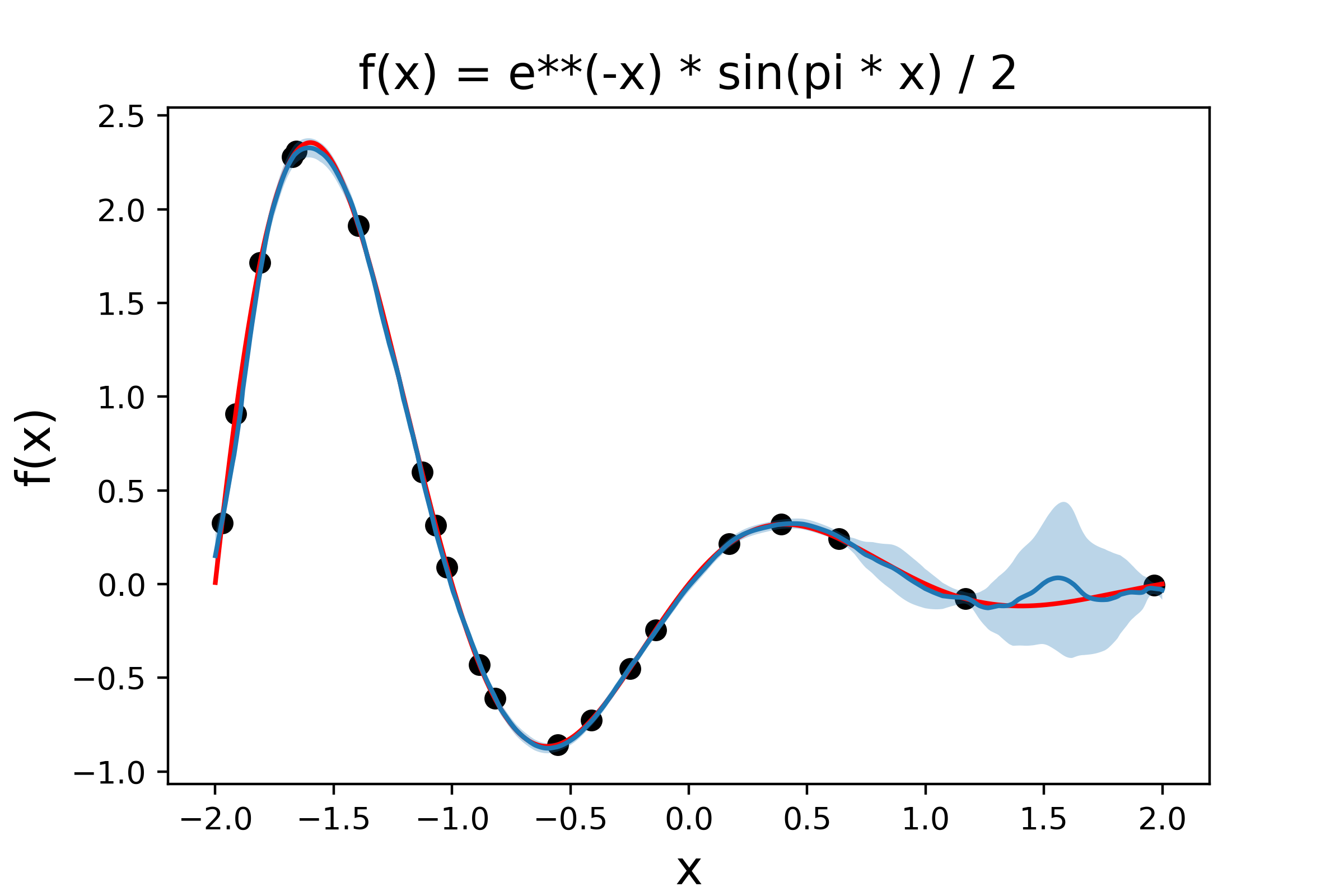}
         \caption{$y=e^{-x} * sin(\pi * x)/2$ }
        \label{fig:1-d_regression:visualization:2}
     \end{subfigure}
     \hfill
     \begin{subfigure}[b]{0.32\textwidth}
         \centering
         \includegraphics[width=\textwidth]{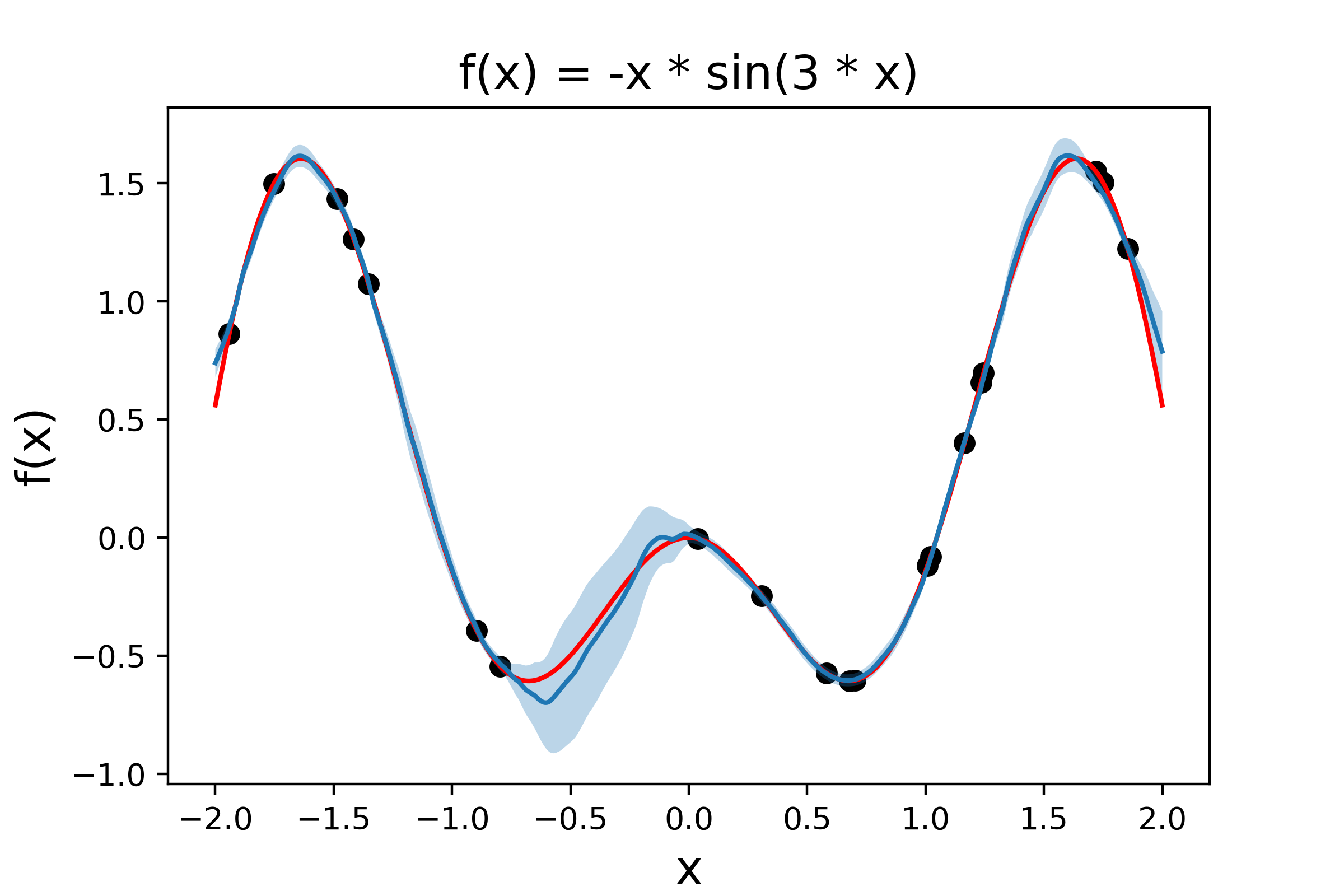}
         \caption{$y=-x * sin(3*x)$}
        \label{fig:1-d_regression:visualization:3}
     \end{subfigure}
    \caption{CMANPs Meta-Regression Visualizations.}
    \label{fig:1-d_regression:visualization}
\end{figure}

\begin{figure}
     \centering

     \begin{subfigure}[b]{0.32\textwidth}
         \centering
         \includegraphics[width=\textwidth]{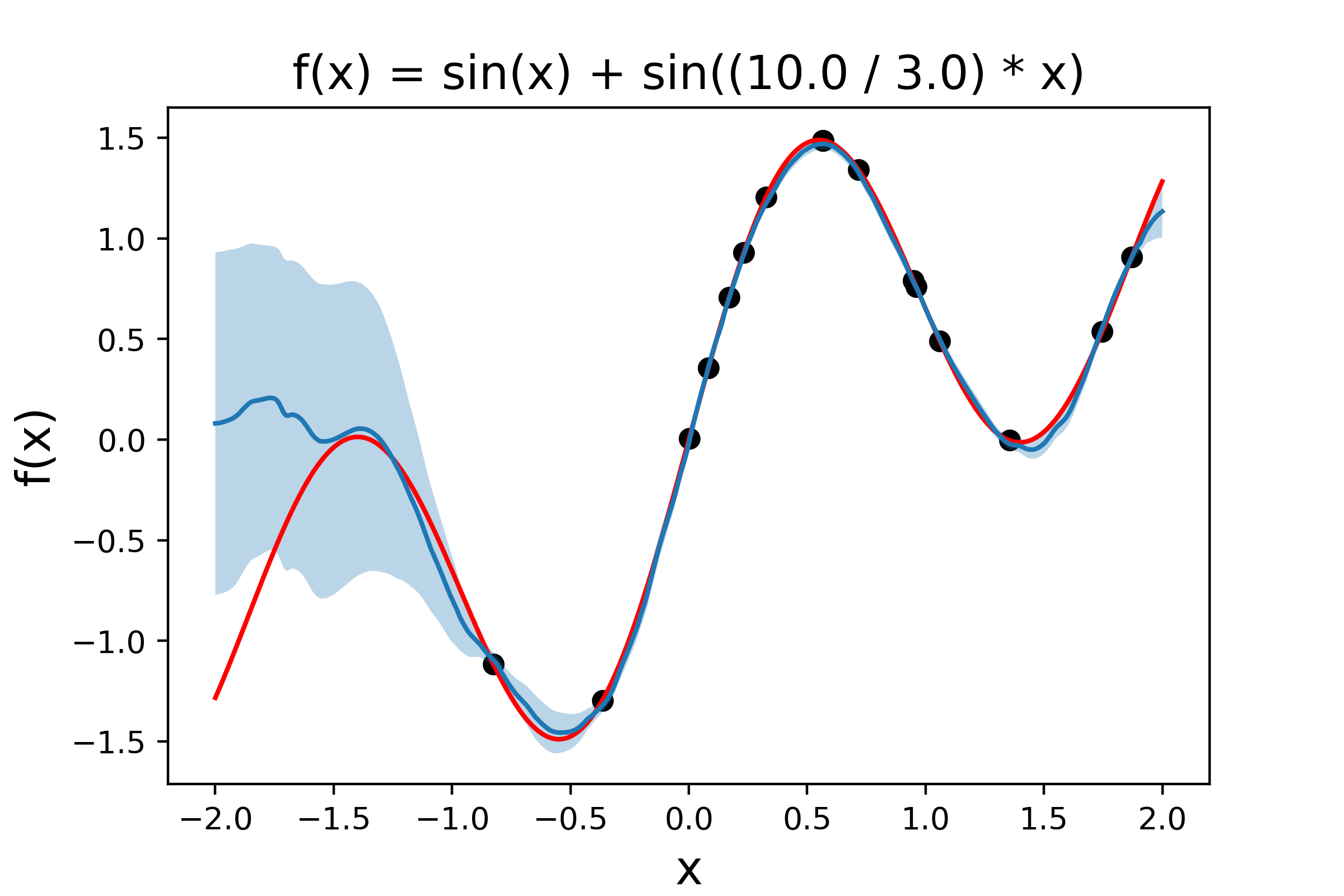}
         \caption{$y=sin(x) + sin(10/3) * x$}
        \label{fig:1-d_regression:ood_visualization:1}
     \end{subfigure}
     \hfill
     \begin{subfigure}[b]{0.32\textwidth}
         \centering
         \includegraphics[width=\textwidth]{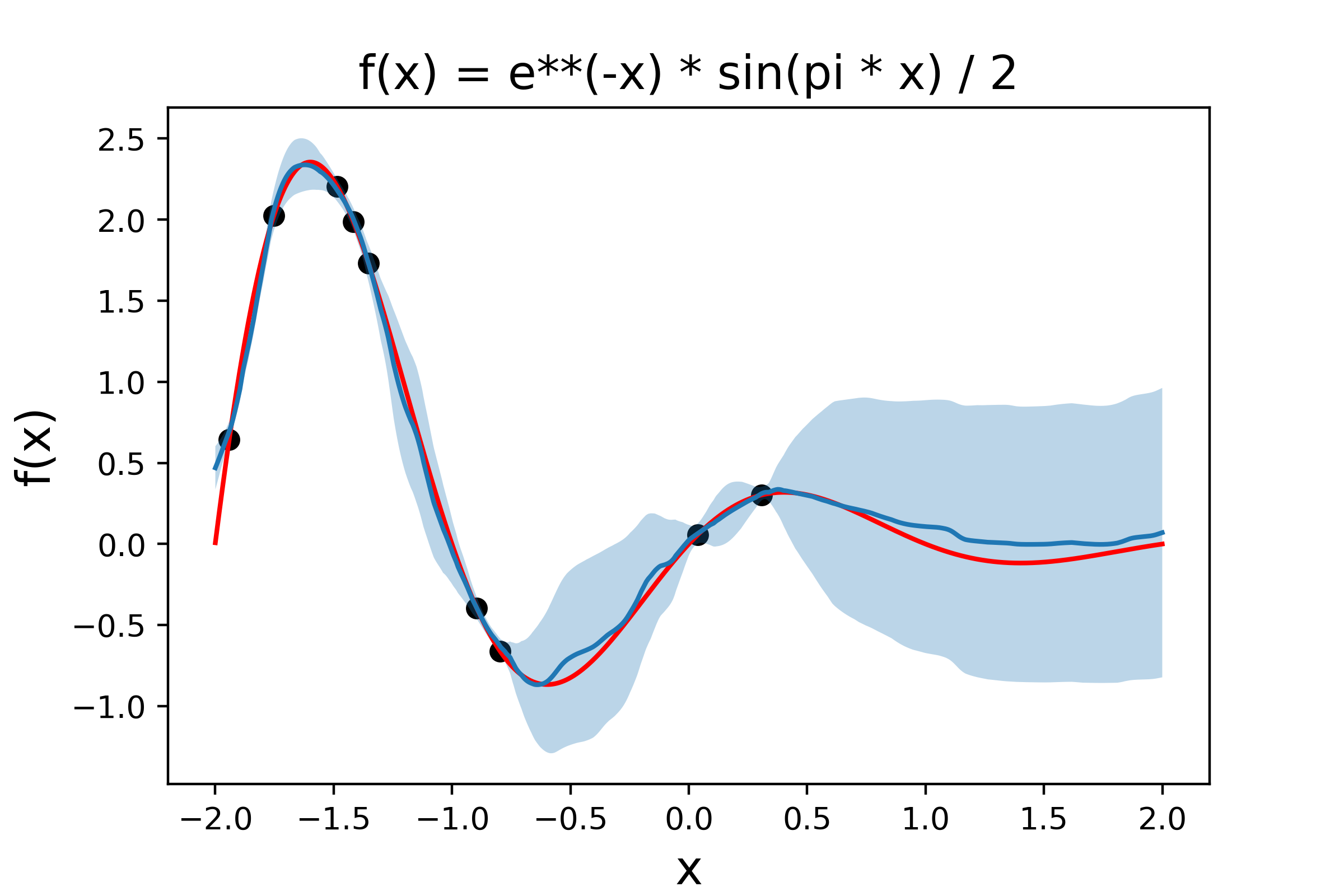}
         \caption{$y=e^{-x} * sin(\pi * x)/2$ }
        \label{fig:1-d_regression:ood_visualization:2}
     \end{subfigure}
     \hfill
     \begin{subfigure}[b]{0.32\textwidth}
         \centering
         \includegraphics[width=\textwidth]{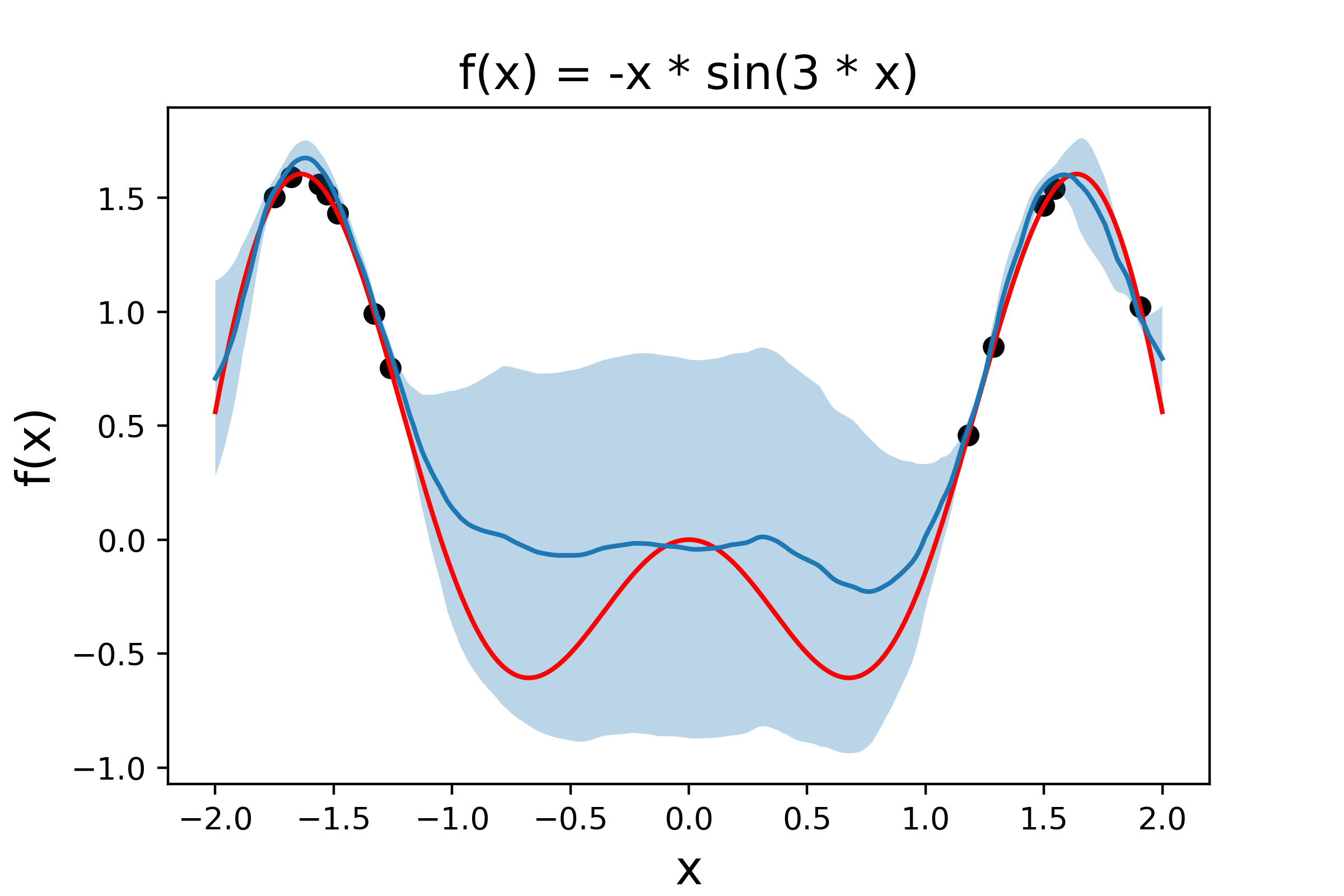}
         \caption{$y=-x * sin(3*x)$}
        \label{fig:1-d_regression:ood_visualization:3}
     \end{subfigure}
    \caption{CMANPs Meta-Regression Out-of-Distribution Visualizations. The model is evaluated between $[-2.0, 2.0]$. However, context data points are sampled from only (a) $[-1.0, 2.0]$, (b) $[-2.0, 1.0]$, and (c) $[-2.0, -1.0] \cup [1.0, 2.0]$.}
    \label{fig:1-d_regression:ood_visualization}
\end{figure}

\begin{figure*}
    \centering
    \includegraphics[width=0.7\textwidth]{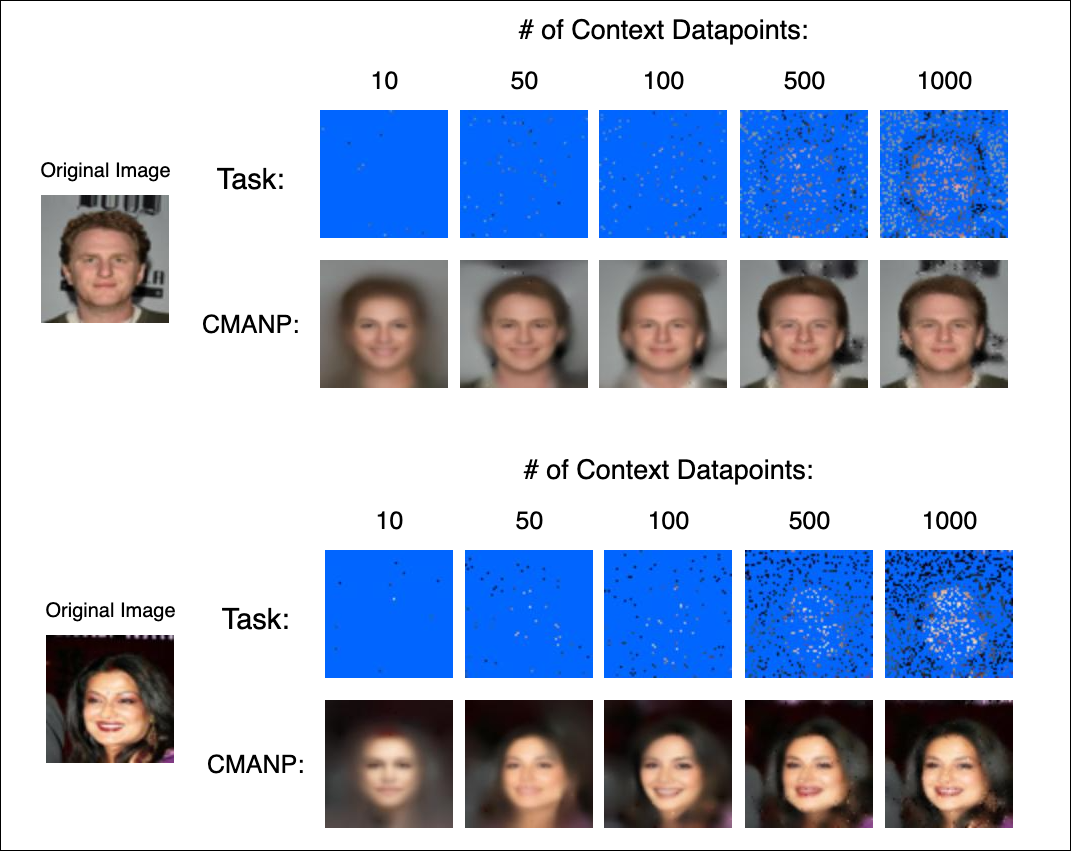}
    \caption{CMANPs Image Completion Visualizations}
    \label{fig:image-completion:visualization}
\end{figure*}

\textbf{Visualizations:} In Figure \ref{fig:1-d_regression:visualization}, we show visualizations for the Meta-Regression task. In Figure \ref{fig:1-d_regression:ood_visualization}, we show out-of-distribution visualizations for the Meta-Regression task. In Figure \ref{fig:image-completion:visualization}, we show visualizations for the Image Completion task.

\section{Appendix: Discussion}

\subsection{Likelihood computation of Autoregressive Not-Diagonal extension compared with that of Not-Diagonal extension:} In the Autoregressive Not-Diagonal extension the predictions are made autoregressively, allowing for the modelling of more flexible distributions than prior Not-Diagonal variants.  As such, the autoregressive not-diagonal variant's likelihood is typically higher than that of the non-autoregressive baselines which only model an unimodal gaussian distribution. 

Consider the following didactic example where $B_Q = 1$ (the block prediction size). Since -AND feeds earlier samples back into the model for making predictions, the likelihood of the target data points: $\{(x_i, y_i)\}^{M}_{i=1}$ for our -AND model is computed as follows:

\begin{align*}
    \log p_{AND}(y_{1:M} | x_{1:M}, \mathcal{D}_C) &= \log \prod_{i=1}^M p(y_{i} | x_{1:i-1}, y_{1:i-1}, x_{i}, \mathcal{D}_C) \\ 
    &= \sum_{i=1}^M \log p(y_{i} | x_{1:i-1}, y_{1:i-1}, x_{i}, \mathcal{D}_C)
\end{align*}
In contrast, consider the likelihood of -ND: $\log p_{ND}(y_{1:M} | x_{1:M}, \mathcal{D}_C)$. 
By Boole's Inequality (or Union Bound), we have that 

$$
    \log p_{ND}(y_{1:M} | x_{1:M}, \mathcal{D}_C) \leq \sum_{i=1}^{M} \log p(y_i | x_{1:M}, \mathcal{D}_C) = \sum_{i=1}^{M} \log p(y_i | x_i, \mathcal{D}_{C})
$$

$(x_{1:i-1}, y_{1:i-1})$ provides relevant information for predicting the value of the function at $x_i$, e.g., nearby pixel values in image completion. As a result, it is likely the case that:

$$
    p(y_i | x_i, \mathcal{D}_C) \leq p(y_{i} | x_{1:i-1}, y_{1:i-1}, x_{i}, \mathcal{D}_C) 
$$
Summing from $i=1 \ldots M$, this means:

$$
    \log p_{ND}(y_{1:M} | x_{1:M}, \mathcal{D}_C) \leq \log p_{AND}(y_{1:M} | x_{1:M}, \mathcal{D}_C)
$$

As such, the Autoregressive Not-Diagonal variant's likelihood is typically higher than that of the Not-Diagonal baselines (i.e., non-autoregressive variants).

\section{Appendix: Implementation, Hyperparameter Details, and Compute}

\subsection{Implementation and Hyperparameter Details}
We use the implementation of the baselines from the official repository of TNPs (\href{https://github.com/tung-nd/TNP-pytorch}{https://github.com/tung-nd/TNP-pytorch}) and LBANPs (\href{https://github.com/BorealisAI/latent-bottlenecked-anp}{https://github.com/BorealisAI/latent-bottlenecked-anp}). The datasets are standard for Neural Processes and are available in the same link.
We follow closely the hyperparameters of TNPs and LBANPs.
In CMANP, the number of blocks for the conditioning phase is equivalent to the number of blocks in the conditioning phase of LBANP.
Similarly, the number of cross-attention blocks for the querying phase is equivalent to that of LBANP.
We used an ADAM optimizer with a standard learning rate of $5e-4$. We performed a grid search over the weight decay term $\{0.0, 0.00001, 0.0001, 0.001\}$. 
Consistent with prior work~\citep{feng2023latent} who set their number of latents $L=128$, we also set the number of latents to the same fixed value $L_I = L_B = 128$ without tuning. 
Due to CMANPs and CMABs architecture, they allow for varying embedding sizes for the learned latent values ($L_I$ and $L_B$). 
For simplicity, we set the embedding sizes to $64$ consistent with prior works~\citep{nguyen2022transformer,feng2023latent}. 
The block size for CMANP-AND is set as $b_Q = 5$. 
During training, CelebA (128x128), (64x64), and (32x32) used a mini-batch size of 25, 50, and 100 respectively. All experiments are run with $5$ seeds.
For the Autoregressive Not-Diagonal experiments, we follow TNP-ND and LBANP-ND~\citep{nguyen2022transformer,feng2023latent} and use cholesky decomposition for our LBANP-AND experiments.
Focusing on the efficiency aspect, we follow LBANPs in the experiments and consider the conditional variant of NPs, optimizing the log-likelihood directly.  

\subsection{Compute}

All experiments were run on a Nvidia GTX 1080 Ti (12 GB) or Nvidia Tesla P100 (16 GB) GPU.
Meta-regression experiments took 4 hours to train.
EMNIST took 2 hours to train. 
CelebA (32x32) took 16 hours to train.
CelebA (64x64) took 2 days to train.
CelebA (128x128) took 3 days to train.

\end{document}